\documentclass[10pt,journal,compsoc]{IEEEtran}
\ifCLASSOPTIONcompsoc
  \usepackage[nocompress]{cite}
\else
  \usepackage{cite}
\fi

\usepackage{times}
\usepackage{epsfig}
\usepackage{graphicx}
\usepackage{amsmath}
\usepackage{amssymb}
\usepackage{subfig}
\usepackage{url}

\begin{document}
\title{Hierarchical Deep Temporal Models for Group Activity Recognition}

\author
{
 Mostafa S. Ibrahim, Srikanth Muralidharan, Zhiwei Deng, Arash Vahdat, Greg Mori,~\IEEEmembership{Member,~IEEE}

 \IEEEcompsocitemizethanks
 {
 \IEEEcompsocthanksitem M. S. Ibrahim, S. Muralidharan., Z. Deng, A. Vahdat, and G. Mori are with the School of Computing Science, Simon Fraser University, Burnaby, BC, V5A 1S6, Canada.
 \protect\\
 E-mail: {msibrahi, smuralid, zhiweid, avahdat}@sfu.ca, mori@cs.sfu.ca.
 }
}

\markboth{IEEE TRANSACTIONS ON PATTERN ANALYSIS AND MACHINE INTELLIGENCE,  VOL.,  NO.}%
{Shell \MakeLowercase{\textit{et al.}}: Bare Demo of IEEEtran.cls for Computer Society Journals}

\IEEEtitleabstractindextext{%
\begin{abstract}
   In this paper we present an approach for classifying the activity performed by a group of people in a video sequence.  This problem of group activity recognition can be addressed by examining individual person actions and their relations.  Temporal dynamics exist both at the level of individual person actions as well as at the level of group activity.  Given a video sequence as input, methods can be developed to capture these dynamics at both person-level and group-level detail.  We build a deep model to capture these dynamics based on LSTM (long short-term memory) models. In order to model both person-level and group-level dynamics, we present a 2-stage deep temporal model for the group activity recognition problem. In our approach, one LSTM model is designed to represent action dynamics of individual people in a video sequence and another LSTM model is designed to aggregate person-level information for group activity recognition.  We collected a new dataset consisting of volleyball videos labeled with individual and group activities in order to evaluate our method.  Experimental results on this new Volleyball Dataset and the standard benchmark Collective Activity Dataset demonstrate the efficacy of the proposed models.
\end{abstract}

}

\maketitle

\IEEEdisplaynontitleabstractindextext
\IEEEpeerreviewmaketitle

\IEEEraisesectionheading{\section{Introduction}\label{sec:introduction}}

\begin{figure}[t]
 \includegraphics[width=1.0\linewidth]{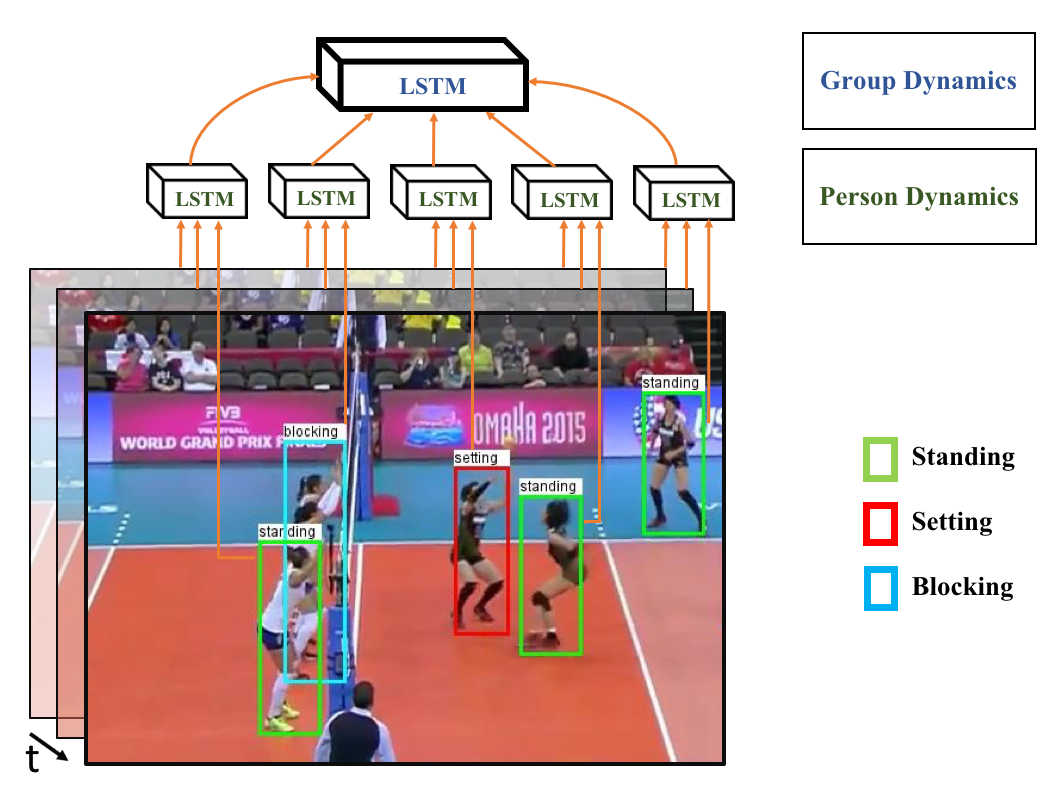}
 \caption{Group activity recognition via a hierarchical model. Each person in a scene is modeled using a temporal model that captures his/her dynamics. These models are integrated into a higher-level model that captures scene-level group activity.}
 \label{fig_illus}
 \end{figure}

\IEEEPARstart
% Modified on 6-4-16
 {W}{E} could describe the action that is happening in Figure~\ref{fig_illus} in numerous levels of abstraction. For instance, we could describe the scene in terms of what each individual player is doing. This task of person-level action recognition is an important component of visual understanding.  At another level of detail, we could instead ask what is the overarching group activity that is depicted.  For example, this frame could be labeled as ``right team setting."  In this paper, we focus on this higher-level group activity task, devising methods for classifying a video according to the activity that is being performed by the group as a whole.

% Added on 6-4-16
 Human activity recognition is a challenging computer vision problem and has received a lot of attention from the research community. It is a challenging problem due to factors such as the variability within action classes, background clutter, and similarity between different action classes, to name a few. Group activity recognition finds a lot of applications in the context of video surveillance, sport analytics, video search and retrieval.  A particular challenge of group activity recognition is the fact that the inference of the label for a scene can be quite sensitive to context.  For example, in the volleyball scene shown in Fig.~\ref{fig_illus}, the group activity hinges on the action of one key individual who is performing the ``setting" action -- though other people in the scene certainly provide helpful information to resolve ambiguity.  In contrast, for group activity categories such as ``talking" or ``queuing" (e.g.\ Fig.~\ref{fig:vis_cad}), the group activity label depends on the actions of many inter-related people in a scene.  As such, successful models likely require the ability to aggregate information across the many people present in a scene and make distinctions utilizing all of this information.

 Spatio-temporal relations among the people in the scene have been at the crux of several approaches in the past that dealt with group activity recognition. The literature shows that spatio-temporal appearance/motion properties of an individual  and their relations can discern which group activity is present.  A volume of research has explored models for this type of reasoning~\cite{choi2009they,lan2012social,ramanathan2013social,amer2014hirf}.  
 These approaches that utilize underlying person-level action recognition based on hand-crafted feature representations including histogram of gradients (HOG) or motion boundary histograms (MBH) both in a dense and sparse fashion~\cite{wang2011action},~\cite{schuldt2004recognizing}.  However, since they rely on shallow hand crafted feature representation, they are limited by their representational abilities to model a complex learning task.  Similarly, the higher-level group activity recognition models utilize on probabilistic or discriminative models built from relatively limited components. 
 
On the other hand, deep representations have overcome this limitation and yielded state of the art results in several computer vision benchmarks~\cite{simonyan2014two},~\cite{karpathy2014large},~\cite{krizhevsky2012imagenet}.
A naive approach to group activity recognition with a deep model would be to simply treat an image as an holistic input.  One could train a model to classify this image according to the group activity taking place. However, it isn't clear if this will work given the redundancy in the training data: with a dataset of volleyball videos, frames will be dominated by features of volleyball courts. As a result, we might have dominant set of features from the non-discriminative, uninteresting regions in the frame, that is common to multiple classes. This might result in poorer performance of the classifier.

  The inter-class distinctions in group activity recogntion arise from the variations in spatio-temporal relations between people, beyond just global appearance. Utilizing a deep model to learn invariance to translation, to focus on the relations between people, presents a significant challenge to the learning algorithm. Similar challenges exist in the object recognition literature, and research often focuses on designing pooling operators for deep networks (e.g.~\cite{Szegedy15}) that enable the network to learn effective classifiers.
  
  % Added on 5-4-2016.
  Group activity recognition presents a similar challenge -- appropriate networks need to be designed that allow the learning algorithm to focus on differentiating higher-level classes of activities. A simple solution to come up with such a representation is to have a layered approach in which each layer focuses on a subset of the image, and a given layer collects the information learnt from its previous layer to learn the higher level information. Hence, we develop a novel hierarchical deep temporal model. This consists of one dedicated layer which reasons about individual people and a second higher level layer that collects the information from the previous layer and learns discriminative frame level information for group activity recognition.
  
  % Modified on 5-4-2016.
  Our method starts with a set of detected and tracked people.
  Given a set of detected and tracked people, we use temporal deep networks (LSTMs) to analyze each individual person. These person-level LSTMs are aggregated over the people in a scene into a higher level deep temporal model. This allows the deep model to learn the relations between the people (and their appearances) that contribute to recognizing a particular group activity. Through this work we show that we can use LSTMs as a plausible deep learning alternative to the graphical models previously used for this task.
  
  The contribution of this paper is the this novel deep architecture that models group activities in a principled structured temporal framework. Our 2-stage approach models individual person activities in its first stage, and then combines person-level information to represent group activities. The model's temporal representation is based on the long short-term memory (LSTM): recurrent neural networks such as these have recently demonstrated successful results in sequential tasks such as image captioning \cite{donahue2014long} and speech recognition
  \cite{graves2014towards}. 
  Through the model structure, we aim at constructing a representation that leverages the discriminative information in the hierarchical structure between individual person actions and group activities. 
  
  We show that our algorithm works in two scenarios. First, we demonstrate performance on the Collective Activity Dataset~\cite{choi2009they}, a surveillance-type video dataset. We also propose a new Volleyball Dataset that offers person detections, and both the person action label as well as the group activity label.  Experimentally, the model is effective in recognizing the overall team activity based on recognizing and integrating player actions. 
  
  This paper builds upon a previous version of this work~\cite{msibrahi16deepactivity}. Here, we present a modified model for alternative pooling structures, an enlarged Volleyball Dataset, and additional empirical evaluations and analyses.

  This paper is organized as follows. In Section~\ref{sec:related_work}, we provide a brief overview of the literature related to activity recognition. In Section~\ref{sec:model}, we elaborate details of the proposed group activity recognition model. In Section~\ref{sec:experiments}, we tabulate the performance of the approach, and end in Section~\ref{sec:conclusion} with a conclusion of this work.

\section{Related Work}
\label{sec:related_work}
  Human activity recognition is an active area of research, with many existing algorithms. Surveys by Weinland et al.~\cite{weinland2011survey} and Poppe~\cite{poppe2010survey} explore the vast literature in activity recognition. Here, we will focus on the group activity recognition problem and recent related advances in deep learning.

\subsection{Group Activity Recognition}
  Group activity recognition has attracted a large body of work recently. Most previous work has used hand-crafted features fed to structured models that represent information between individuals in space and/or time domains. For example, Choi et al.~\cite{choi2009they} craft spatio-temporal feature representations of relative human actions.  Lan et al.~\cite{LanWYRM12} proposed an adaptive latent structure learning that represents hierarchical relationships ranging from lower person-level information to higher group-level interactions. 
  
  Lan et al.~\cite{LanSM12} and Ramanathan et al.~\cite{ramanathan2013social} explore the idea of social roles, the expected behaviour of an individual person in the context of group, in fully supervised and weakly supervised frameworks respectively.  Lan et al.~\cite{LanSM12} map the features defined on individuals to group activity by constructing a hierarchical model consisting of individual action, role based unary components, pairwise roles, and scene level group activities. The interactions and unary roles/activities are represented using an undirected graphical model. The parameters of this model are learnt using a structured SVM formulation in a max margin framework, and operates under completely supervised settings.
  
  Ramanathan et al.~\cite{ramanathan2013social} define a CRF-based social role model under a weakly supervised setting. To learn model parameters and role labels, a joint variational inference procedure is adapted. HOG3D~\cite{klaser2008spatio}, spatio-temporal features~\cite{wang2011action}, object interaction feature~\cite{li2010object}, and social role features~\cite{zhu2012face} are used as unary component representations. A subsequent layer consisting of pairwise spatio-temporal interaction features is used to refine the noisy unary component features. Finally, variational inference is used to learn the unknown role labels and model parameters. 
  
  Choi and Savarese~\cite{choiS12} unified tracking multiple people, recognizing individual actions, interactions and collective activities in a joint framework. The model is based on the premise that strong correlation exists between an individual’s activity, and the activities of the other nearby people. Following this intuition, they come up with a hierarchical structure of activity types that maps the individual activity to overall group activity. In this process, they simultaneously track atomic activities, interactions and overall group activities. The parameters of this model (and the inference) are learnt by combining belief propagation with the branch and bound algorithm. 
  
  Chang et al.~\cite{chang2011probabilistic} employ a probabilistic grouping strategy to perform high level recognition tasks happening in the scene. Specifically, group structure is determined by soft grouping structures to facilitate the representation of dynamics present in the scene. Secondly, they also use a probabilistic motion analysis to extract interesting spatio-temporal patterns for scenario recognition.
  Vascon et al.~\cite{VasconZCHPM14} detect conversational groups in crowded scenes of people.  The approach uses pairwise affinities between people based on pose and a game-theoretic clustering procedure.
  
  In other work~\cite{choi2011learning}, a random forest structure is used to sample discriminative spatio-temporal regions from input video fed to 3D Markov random field to localize collective activities in a scene. Shu et al.~\cite{ShuXRTZ15} detect group activities from aerial video using an AND-OR graph formalism. The above-mentioned methods use shallow hand crafted features, and typically adopt a linear model that suffers from representational limitations.

  \subsection{Sport Video Analysis}
  
  Computer vision-based analysis of sports video is a burgeoning area for research, with many recent papers and workshops focused on this topic.  Work on sports video analysis has spanned a range of topics from individual player detection, tracking, and action recognition, to player-player interactions, to team-level activity classifications. Much work spans many of these taxonomy elements, including the seminal work of Intille and Bobick~\cite{IntilleB01}, who examined stochastic representations of American football plays. 
  
  {\bf Player tracking}: Nillius et al.~\cite{nillius2006multi} link player trajectories to maintain identities via reasoning in a Bayesian network formulation.  Morariu et al.~\cite{morariu11eventstructure} track players, infer part locations, and reason about temporal structure in 1-on-1 basketball games. In Soomro et al.~\cite{soomro2015tracking}, a graph based optimization technique is applied to address the task of tracking in broadcast soccer videos where a disjoint temporal sequence of soccer videos is present. They first extract panoramic view video clips, and subsequently detect and track multiple players by a two step bipartite matching algorithm. Bo et al.~\cite{Bo_2013_CVPR_Workshops} introduced a novel approach to scale and rotation invariant tracking of human body parts. They use a dynamic programming based approach that optimizes the assembly of body part region proposals, given spatio-temporal constraints under a loopy body part graph construction, to enable scale and rotation invariance.  
  
  {\bf Actions and player roles:} Turchini et al.~\cite{Turchini_2015_ICCV_Workshops} perform activity recognition by first obtaining dense trajectories~\cite{wang2011action}, clustering them, and finally employ a cluster set kernel to learn a action representations. Kwak et al.~\cite{KwakHH13} optimize based on a rule-based depiction of interactions between people.

  Wei et al.~\cite{Wei_2015_ICCV_Workshops} compute a role ordered feature representation to predict the ball owner at each time instance in a given video. They start from the annotated positions of each player, permute them and obtain the feature representation ordered by relative position (called as role) with respect to other players.
  
  {\bf Team activities:} Siddiquie et al.~\cite{siddiquie2009recognizing} proposed sparse multiple kernel learning to select features incorporated in a spatio-temporal pyramid. In Bialkowski et al.~\cite{bialkowski2013recognising}, two detection based representations that are based on team occupancy map and team centroid map respectively, are shown to effectively detect team activities in field hockey videos. First, players are detected in each of the eight camera views that are used, and then team level aggregations are computed after classifying each player into one of the two teams. Finally, using these aggregated representations, team activity labels are computed.
  
  Atmosukarto et al.~\cite{Atmosukarto_2013_CVPR_Workshops} define an automated approach for recognizing offensive team formation in American football. First, the frame pertaining to the offensive team formation is first identified, line of scrimmage is obtained, and eventually the team formation label is obtained by learning a SVM classifier on top of the offensive team side's features inferred using the line of scrimmage. Direkoglu and O'Connor \cite{direkoǧluO12} solved a Poisson equation to generate a holistic player location representation.   Swears et al.~\cite{SwearsHJB14} used the Granger Causality statistic to automatically constrain the temporal links of a Dynamic Bayesian Network (DBN) for handball videos. 
  
  In Gade et al.~\cite{Gade_2013_CVPR_Workshops}, player occupancy heat maps are employed to handle sport type classification. People are first detected, and the occupancy maps are obtained by summing their locations over time. Finally, a sport type classifier is trained on top of Fisher vector representations of the heat maps to infer the sports type happening in a test scene.

 % commented by mostafa to stop latex errors
% CVPR 2013 - http://www.vap.aau.dk/cvsports/ (computer vision in sports workshop)

%           - http://www.dcs.gla.ac.uk/~vincia/sism2013/index.html (surveillance workshop)
 
% ECCV 2012 - http://www.artemis2012.tuc.gr/ (surveillance)
% CVPR 2012 - http://www.dcs.gla.ac.uk/~vincia/sism/index.html (surveillance)
% CVPR 2011 - http://www.ee.oulu.fi/~gyzhao/MLVMA09/index_files/Page496.htm (surveillance)

  \subsection{Deep Learning}
  Deep Convolutional Neural Networks (CNNs) have shown impressive performance by unifying feature and classifier learning, enabled by the availability of large labeled training datasets. Successes have been demonstrated on a variety of computer vision tasks including image classification~\cite{krizhevsky2012imagenet, simonyan2014very} and action recognition~\cite{simonyan2014two, karpathy2014large}. More flexible recurrent neural network (RNN) based models are used for handling variable length space-time inputs. Specifically, LSTM~\cite{hochreiter1997long} models are popular among RNN models due to the tractable learning framework that they offer when it comes to deep representations. These LSTM models have been applied to a variety of tasks \cite{donahue2014long,graves2014towards,ng2015beyond,venugopalan2014translating}.
  
  For instance, in Donahue et al.~\cite{donahue2014long}, the so-called Long term Recurrent Convolutional network, formed by stacking an LSTM on top of pre-trained CNNs, is proposed for handling sequential tasks such as activity recognition, image description, and video description. In this work, they showed that it is possible to jointly train LSTMs along with convolutional networks and achieve comparable results to the state of the art for time-varying tasks. For example, in video captioning, they first construct a semantic representation of the video using maximum a posteriori estimation of a conditional random field.  This is then used to construct a natural sentence using LSTMs.
  
  In Karpathy et al.~\cite{karpathy2014deep}, structured objectives are used to align CNNs over image regions and bi-directional RNNs over sentences. A deep multi-modal RNN architecture is used for generating image descriptions using the deduced alignments. In the first stage, words and image regions are embedded onto an alignment space. Image regions are represented by RCNN embeddings, and words are represented using bi-directional recurrent neural network~\cite{schuster1997bidirectional} embeddings. In the second stage, using the image regions and textual snippets, or full image and sentence descriptions, a generative model based on an RNN is constructed, that outputs a probability map of the next word.
  
  In this work, we aim at building a hierarchical structured model that incorporates a deep LSTM framework to recognize individual actions and group activities. Previous work in the area of deep structured learning includes Tompson et al.~\cite{NIPS2014_5573} for pose estimation, and Zheng et al.~\cite{Zheng15} and Schwing et al.~\cite{schwing2015fully} for semantic image segmentation. 
  
  In Deng et al.~\cite{DengZCLMRM15} a similar framework is used for group activity recognition, where a neural network-based hierarchical graphical model refines person action labels and learns to predict the group activity simultaneously. While these methods use neural network-based graphical representations, in our current approach, we leverage LSTM-based temporal modelling to learn discriminative information from time varying sports activity data. In \cite{yeung2015every}, a new dataset is introduced that contains dense multiple labels per frame for underlying action, and a novel Multi-LSTM is used to model the temporal relations between labels present in the dataset.  Ramanathan et al.~\cite{RamanathanHAGMF16} develop LSTM-based methods for analyzing sports videos, using an attention mechanism to determine who is the principal actor in a scene.  In a sense, this work is complementary to our pooling-based models that represent aggregations of all people involved in a group activity.

\subsection{Datasets}
  
Popular datasets for activity recognition include the Sports-1M dataset~\cite{karpathy2014deep}, UCF 101 database \cite{soomro2012ucf101}, and the HMDB movie database \cite{kuehne2011hmdb}. These datasets were part of a shift in focus toward unconstrained Internet videos as a domain for action recognition research.  These datasets are challenging because they contain substantial intra-class variation and clutter both in terms of extraneous background objects and varying temporal duration of the action of interest.  However, these datasets tend to focus on individual human actions, as opposed to the group activities we consider in our work.

Scenes involving multiple, potentially interacting people present significant challenges.  In the context of surveillance video, the TRECVid Surveillance Event Detection~\cite{2015trecvidover}, UT-Interaction~\cite{UT-Interaction-Data}, VIRAT~\cite{OhVIRAT11}, and UCLA Courtyard datasets~\cite{amer2012cost} are examples of challenging tasks including individual and pairwise interactions.

Datasets for analyzing group activities include the Collective Activity Dataset~\cite{choi2009they}. This dataset consists of real world pedestrian sequences where the task is to find the high level group activity. 
The S-HOCK dataset~\cite{ConigliaroRSBCSC15} focuses on crowds of spectators and contains more than 100 million annotations ranging from person body poses to actions to social relations among spectators.
In this paper, we experiment with the Collective Activity Dataset, and also introduce a new dataset for group activity recognition in sport footage which is annotated with player pose, location, and group activities\footnote{The dataset is available for download: \url{https://github.com/mostafa-saad/deep-activity-rec}.}.

\section{Proposed Approach}
\label{sec:model}

Our goal in this paper is to recognize activities performed by a group of people in a video sequence. The input to our method is a set of tracklets of the people in a scene.
The group of people in the scene could range from players in a sports video to pedestrians in a surveillance video. In this paper we consider three cues that can aid in determining what a group of people is doing:

\begin{itemize}
\item \textbf{Person-level actions} collectively define a group activity. Person action recognition is a first step toward recognizing group activities.

\item \textbf{Temporal dynamics of a person's action} is higher-order information that can serve as a strong signal for group activity. Knowing how each person's action is changing over time can be used to infer the  group's activity. 

% \item \textbf{Spatial context} can be represented as what most people in the scene are doing at the current time. MostafaComment: We use fc7 over person NOT scene.

\item \textbf{Temporal evolution of group activity} represents how a group's activity is changing over time. For example, in a volleyball game a team may move from defence phase to pass and then attack. 

\end{itemize}

%Popular approaches that address activity recognition use hand crafted feature representation, and a linear model that has limited representational capability \cite{wang2011action} \cite{schuldt2004recognizing}. In the context of group activity recognition, it is shown that representing a group activity using a hierarchical models yield better performance than shallow representations \cite{LanWYRM12, LanSM12, ramanathan2013social}. \\

Many classic approaches to the group activity recognition problem have modeled these elements in a form of structured prediction based on hand crafted features \cite{wang2011action, schuldt2004recognizing, LanWYRM12, LanSM12, ramanathan2013social}. Inspired by the success of deep learning based solutions, in this paper, a novel hierarchical deep learning based model is proposed that is potentially capable of learning low-level image features, person-level actions, their temporal relations, and temporal group dynamics in a unified end-to-end framework.

Given the sequential nature of group activity analysis, our proposed model is based on a Recurrent Neural Network (RNN) architecture. RNNs consist of non-linear units with internal states that can learn dynamic temporal behavior from a sequential input with arbitrary length. Therefore, they overcome the limitation of CNNs that expect constant length input. This makes them widely applicable to video analysis tasks such as activity recognition. 

\begin{figure*}[t]
  \begin{center}
  \includegraphics[trim=0 0 0 0,clip, width=0.9\linewidth]{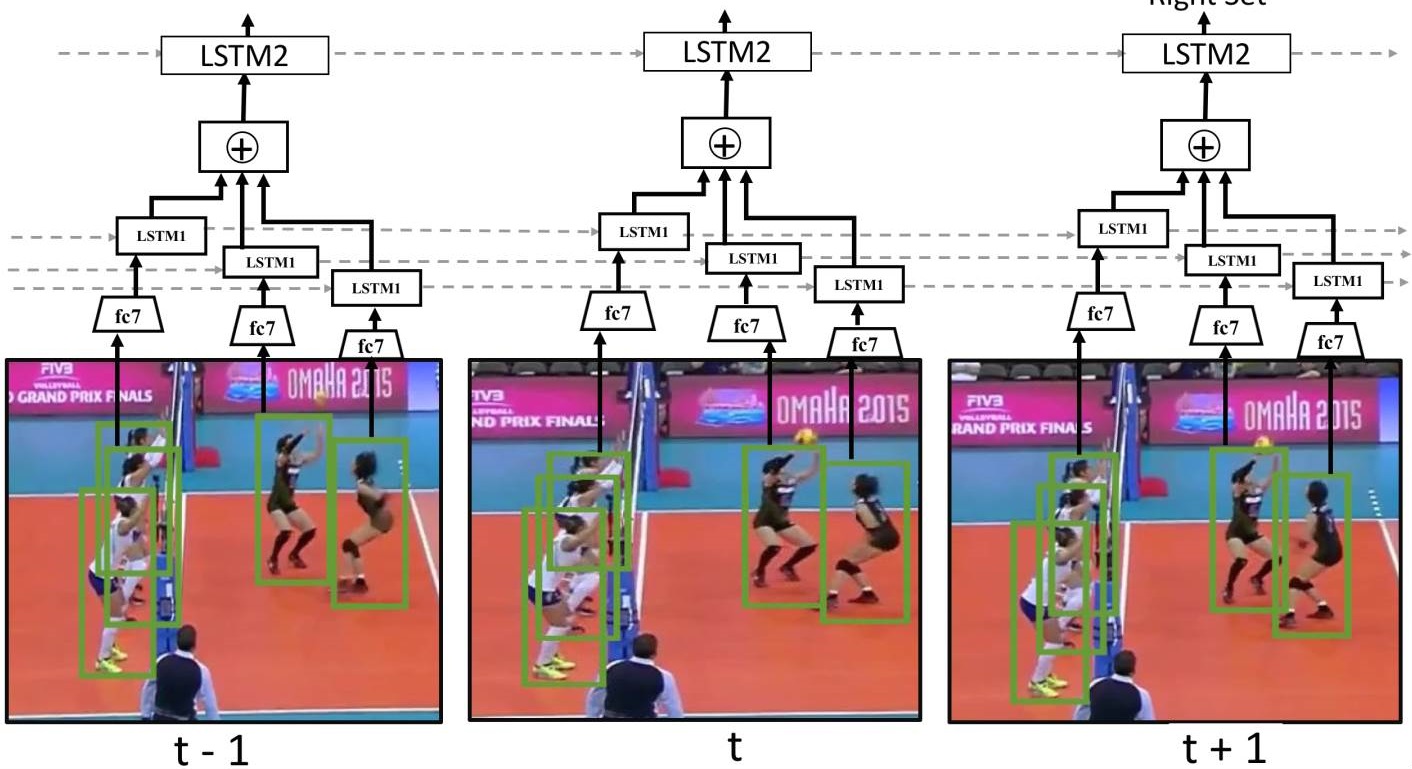}
  \end{center}
    \caption{Our two-stage model for a volleyball match. Given tracklets of K players, we feed each tracklet to a CNN, followed by a person LSTM layer to represent each player's action. We then pool temporal features over all people in the scene. The output of the pooling layer is fed to the second LSTM network to identify the whole team's activity.}
  \label{fig_two_stages_model}
  \end{figure*}

Our model is inspired by the success of hierarchical models.   Here, we aim to mimic a similar intuition using recurrent networks. We propose a deep model by stacking several layers of RNN-type structures to model a range of low-level to high-level dynamics defined on top of people and entire groups. Fig.~\ref{fig_two_stages_model} provides an overview of our model.  We describe the use of these RNN structures for individual and group activity recognition next.

\subsection{Temporal Model of Individual Action} \label{sec:lstm}

Given tracklets of each person in a scene, we use long short-term memory (LSTM) models to represent temporally the action of each individual person. Such temporal information is complementary to spatial features and is critical for performance. LSTMs, originally proposed by Hochreiter and Schmidhuber \cite{hochreiter1997long}, have been used successfully for many sequential problems in computer vision. Each LSTM unit consists of several cells with memory that stores information for a short temporal interval. The memory content of a LSTM makes it suitable for modeling complex temporal relationships that may span a long time range.

The content of the memory cell is regulated by several gating units that control the flow of information in and out of the cells. The control they offer also helps in avoiding spurious gradient updates that can typically happen in training RNNs when the length of a temporal input is large. This property enables us to stack a large number of such layers in order to learn complex dynamics present in the input in different ranges. 

We use a deep Convolutional Neural Network (CNN) to extract features from the bounding box around the person in each time step on a person trajectory. The output of the CNN, represented by $x_t$, can be considered as a complex image-based feature describing the spatial region around a person. Assuming $x_t$ as the input of an LSTM cell at time $t$, the cell activition can be formulated as :
  \begin{flalign}
  i_{t} &= \sigma(W_{xi}x_{t} + W_{hi}h_{t - 1} + b_{i}) \\
  f_{t} &= \sigma(W_{xf}x_{t} + W_{hf}h_{t - 1} + b_{f}) \\
  o_{t} &= \sigma(W_{xo}x_{t} + W_{ho}h_{t - 1} + b_{o}) \\ 
  g_{t} &= \phi(W_{xc}x_{t} + W_{hc}h_{t - 1} + b_{c}) \\
  c_{t} &= f_t \odot c_{t - 1} + i_t \odot g_t \\ 
  h_{t} &= o_t \odot \phi(c_{t})
  \end{flalign}
  
  Here, $\sigma$ stands for a sigmoid function, and $\phi$ stands for the tanh function. $x_t$ is the input, $h_t \in \mathbb{R}^N$ is the hidden state with N hidden units, $c_t \in \mathbb{R}^N$ is the memory cell, $i_t \in \mathbb{R}^N$, $f_t \in \mathbb{R}^N$, $o_t \in \mathbb{R}^N$, and, $g_t \in \mathbb{R}^N$ are input gate, forget gate, output gate, and input modulation gate at time $t$ respectively. $\odot$ represents element-wise multiplication.
  
  When modeling individual actions, the hidden state $h_t$ could be used to model the action a person is performing at time $t$. Note that the cell output is evolving over time based on the past memory content. Due to the deployment of gates on the information flow, the hidden state will be formed based on a short-range memory of the person's past behaviour.
  Therefore, we can simply pass the output of the LSTM cell at each time to a softmax classification layer\footnote{More precisely, a fully connected layer fed to softmax loss layer.} to predict individual person-level action for each tracklet.
  
  The LSTM layer on top of person trajectories forms the first stage of our hierarchical model. This stage is designed to model \textbf{person-level actions and their temporal evolution}. Our training proceeds in a stage-wise fashion, first training to predict person level actions, and then pasing the hidden states of the LSTM layer to the second stage for group activity recognition, as discussed in the next section.
  
  %Our model learns dynamic representations right from primitive actions to group activities using LSTMs. Through this approach, we show that LSTM's can be used simultaneously for feature learning, and also for constructing a discriminative representation of group activities. We next describe the pre-processing steps that we performed before conducting the experiment, and then follow it up with the description of our model.

  %Concretely, a semantic representation that captures different levels of interaction
 % (e.g. person-person interactions) distinctly is likely to be more successful than a model that does not possess such semantic structuring

\subsection{Hierarchical Model for Group Activity Recognition}

%The goal of group activity recognition is to assign a video clip to one of K activity classes. Given a video clip, we detect and track individuals in the scene. 

%We propose a hierarchical LSTM model for group activity recognition as shown in Figure~\ref{fig_two_stages_model}. The first level of our LSTM model is designed to model \textbf{person-level actions and their temporal evolution}. Given a person tracklet, we use a deep Convolutional Neural Network (CNN) to extract features from the bounding box around the person in each time step. The output of the CNN is directly fed to an LSTM layer that captures temporal dynamics for each person. By adding a classification layer (i.e. softmax) on top of the first LSTM layer and using an end-to-end training of both CNN and the LSTM layer, we are able to learn discriminative image-based features for action recognition from individual human tracklets.

At each time step, the memory content of the first LSTM layer contains discriminative information describing the subject's action as well as past changes in his action. If the memory content is  collected over all people in the scene, it can be used to describe the group activity in the whole scene. 

Moreover, it can also be observed that direct image-based features extracted from the spatial domain around a person carry a discriminative signal for the current activity. Therefore, a deep CNN model is used to extract complex features for each person in addition to the temporal features captured by the first LSTM layer. 

At this moment, the concatenation of the CNN features and the LSTM layer represent temporal features for a person. Various pooling strategies can be used to aggregate these features over all people in the scene at each time step. The output of the pooling layer forms our representation for the group activity. The second LSTM network, working on top of the temporal representation, is used to directly model the \textbf{temporal dynamics of group activity}. The LSTM layer of the second network is directly connected to a classification layer in order to detect group activity classes in a video sequence. 

Mathematically, the pooling layer can be expressed as the following:
 \begin{flalign}
  P_{tk} &= x_{tk} \oplus h_{tk} \\
  Z_{t} &= P_{t1} \diamond P_{t2} \ ... \diamond P_{tk}
\end{flalign}

In this equation, $h_{tk}$ corresponds to the first stage LSTM output, and $x_{tk}$ corresponds to the AlexNet fc7 feature, both obtained for the k\textsuperscript{th} person at time t. We concatenate these two features (represented by $\oplus$) to obtain the temporal feature representation $P_{tk}$ for k\textsuperscript{th} person. We then construct the frame level feature representation $Z_{t}$ at time t by applying a max pooling operation (represented by $\diamond$) over the features of all the people. Finally, we feed the frame level representation to our second LSTM stage that operates similar to the person level LSTMs that we described in the previous subsection, and learn the group level dynamics. $Z_{t}$, passed through a fully connected layer, is given to the input of the second-stage LSTM layer.  The hidden state of the LSTM layer represented by $h_{t}^{group}$ carries temporal information for the whole group dynamics. $h_{t}^{group}$ is fed to a softmax classification layer to predict group activities. 
% \begin{flalign}
%  h^{fr}_{t} &= F(h^{fr}_{t - 1}, Z_{t})
% \end{flalign}
 
%Here $h^{fr}_{t}$ corresponds to the representation of t\textsuperscript{th} frame obtained by a transformation $F$ that is identical to the set of transformations explained in the previous subsection.

 %Next, we describe the implementation details that were used for constructing our model.

%The second stage model consists of a single LSTM layer that learns to predict frame level group activity labels. For doing this, we aggregate the person level features to generate frame level features. We try different aggregation techniques for generating the frame level features. Specifically, we consider max-pooling, average pooling, and sum pooling. We tried mimicing operation of cardinality kernel \cite{hajimirsadeghi2015visual} that yielded good performance in the past using our average pooling and sum pooling operation. However, as shown in thit turned out that they were not better than max pooling.

%  To propose people' locations for a foreground blob, we assume a fixed scale (w, h) for the person's bounding box (e.g. the moving camera almost has the same zoom). We use a simple brute force approach that tries every rectangle of scale (w, h) with step size (w/2, h/4) in the (right, bottom) directions assuming we start from the most top left location in the blob rectangle.

%------------------------------------------------------------------------
% New section
\subsection{Handling sub-groups}
\begin{figure}
  \begin{center}
  \includegraphics[trim=0 0 0 0,clip, width=0.9\linewidth]{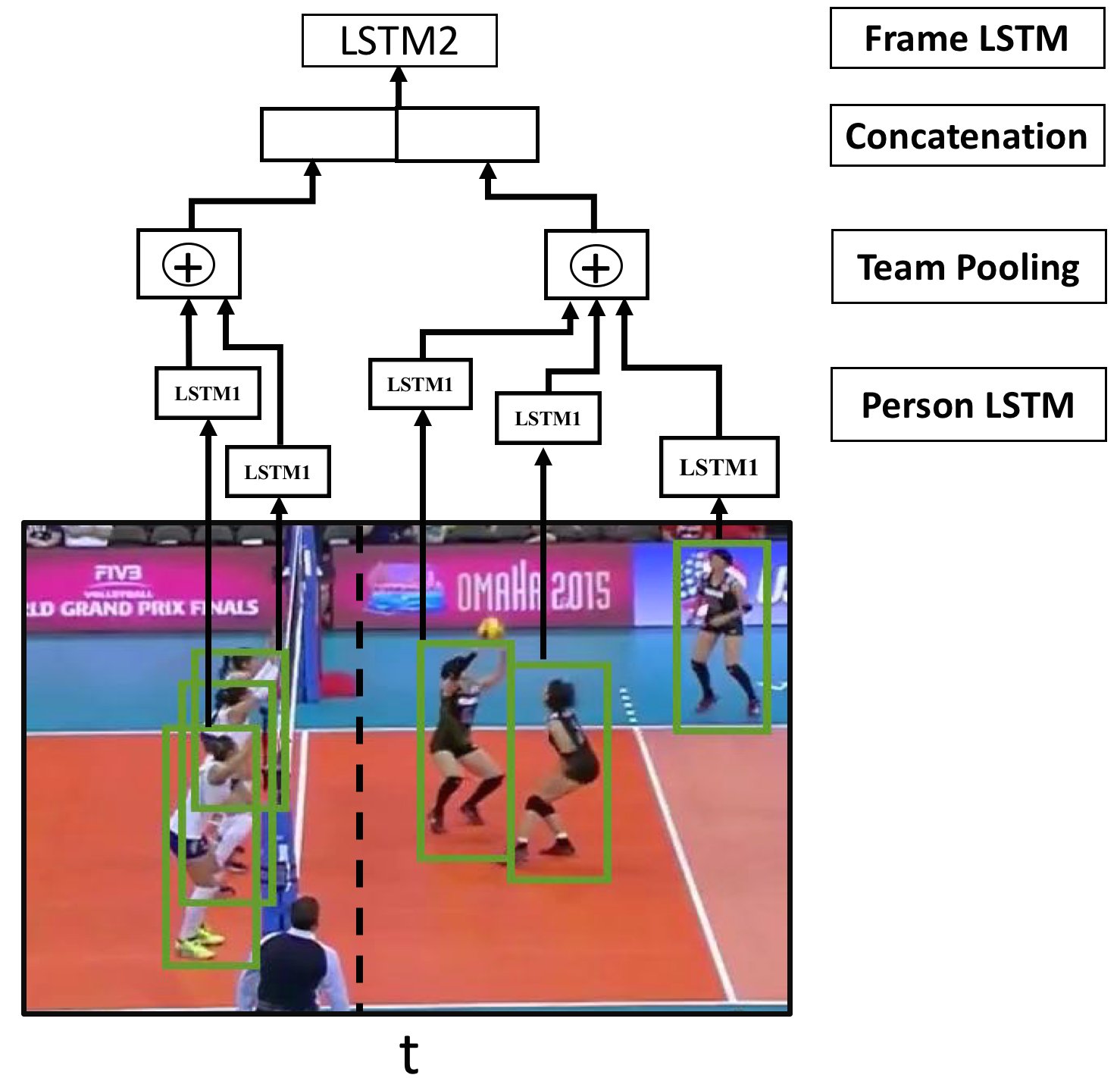}
  \end{center}
    \caption{Illustration of 2-group pooling to capture spatial arrangements of players.}
  \label{fig_two_stages_model_2_blocks}
  \end{figure}

In team sports, there might be several sub-groups of players with common responsibilities within a team. For example, the front players of a volleyball team are responsible for blocking the ball. Max pooling all players' representation in one representation reduces the model capabilities (e.g. causes confusions between left team and right team activities). To consider that, we propose a modified model where we split the players to several sub-groups and recognize the team activity based on the concatenation of each sub-group's representation. In our experiments we consider a set of different possible spatial sub-groupings of players (c.f.\ standard spatial pyramids~\cite{lazebnik2006beyond}). Figure~\ref{fig_two_stages_model_2_blocks} illustrates this variant of the model, showing splitting into two team-based groups.

Mathematically, the pooling layer can be re-expressed as the following:
 \begin{flalign}
  P_{tk} &= x_{tk} \oplus h_{tk} \\
  S_{m} &= (m-1) * k / d + 1 \\
  E_{m} &= m * k / d \\
  G_{tm} &= P_{tSm} \diamond P_{t(Sm+1)} \ ... \diamond P_{tEm} \\
  Z_{t} &= G_{t1} \oplus G_{t2} \ ...  \oplus G_{td}
 \end{flalign}
 
where again, $t$ indexes time, $k$ indexes players, $h_{tk}$ corresponds to first stage LSTM output, $x_{tk}$ to fc7 features, and $P_{tk}$ is the temporal feature representation for the player. Assume that the $K$ players are ordered in a list (e.g. based on top-left point of a bounding box), $d$ is the number of sub-groups and $m$ indexes the groups. $S_{m}$ and $E_{m}$ are the start and end positions of the m-th group players. $G_{tm}$ is the $m$-th group representation: a max pooling on all group players' representation in this group. $Z_{t}$ is the the frame level feature representation constructed by the concatenation operator (represented by $\oplus$) of the $d$ sub-groups.

%------------------------------------------------------------------------
\subsection{Implementation Details}
We trained our model in two steps. In the first step, the person-level CNN and the first LSTM layer are trained in an end-to-end fashion using a set of training data consisting of person tracklets annotated with action labels. We implement our model using Caffe~\cite{Caffe}. Similar to other approaches \cite{donahue2014long, DengZCLMRM15, venugopalan2014translating}, we initialize our CNN model with the pre-trained AlexNet network and we fine-tune the whole network for the first LSTM layer. 

After training the first LSTM layer, we concatenate the fc7 layer of AlexNet and the LSTM layer for every person and pool over all people in a scene. The pooled features, which correspond to frame level features, are fed to the second LSTM network.

For training all our models, we follow the same training protocol. We use a fixed learning rate of 0.00001 and a momentum of 0.9. For tracking subjects in a scene, we used the tracker by Danelljan et al.~\cite{Danelljan14_tracker}, implemented in the Dlib library~\cite{dlib09}. The baseline models are structured and trained in a similar manner as our two-stage model.

\section{Experiments}
\label{sec:experiments}
In this section, we evaluate our model by running ablation studies using several baselines and comparing to previously published works on the Collective Activity Dataset~\cite{choi2009they}.
First, we describe our baseline models for the ablation studies. Then, we present our results on the Collective Activity Dataset followed by experiments on the Volleyball Dataset.

\subsection{Baselines}
The following baselines are considered in all our experiments in order to assess the contributions of components of our proposed model.

{\renewcommand{\labelenumi}{B\theenumi)}
\begin{enumerate}
\item \textbf{Image Classification: } %B1
This baseline is the basic AlexNet model fine-tuned for group activity recognition in a single frame.

\item \textbf{Person Classification:} % B4
In this baseline, the AlexNet CNN model is deployed on each person, fc7 features are pooled over all people, and are fed to a softmax classifier to recognize group activities in each single frame. %The whole network is fine-tuned from the pre-trained AlexNet model.

\item \textbf{Fine-tuned Person Classification:} %B4'
This baseline is similar to the previous baseline with one distinction. The AlexNet model on each player is fine-tuned to recognize person-level actions. Then, fc7 is pooled over all players to recognize group activities in a scene without any fine-tuning of the AlexNet model. 

The rationale behind this baseline is to examine a scenario where person-level action annotations as well as group activity annotations are used in a deep learning model that does not model the temporal aspect of group activities. This is very similar to our two-stage model without the temporal modeling.

%that learning is done in two stages. First, action-level classifier is trained and then features representing players are used to train group-activities.

\item \textbf{Temporal Model with Image Features:} %B2
This baseline is a temporal extension of the first baseline. It examines the idea of feeding image level features directly to a LSTM model to recognize group activities. In this baseline, the AlexNet model is deployed on the whole image and resulting fc7 features are fed to a LSTM model. This baseline can be considered as a reimplementation of Donahue et al.~\cite{donahue2014long}.

\item \textbf{Temporal Model with Person Features:} %B3
This baseline is a temporal extension of the second baseline: fc7 features pooled over all people are fed to a LSTM model to recognize group activities.

\item \textbf{Two-stage Model without LSTM 1:} %B3
This baseline is a variant of our model, omitting the person-level temporal model (LSTM 1). Instead, the person-level classification is done only with the fine-tuned person CNN.

\item \textbf{Two-stage Model without LSTM 2:} %B3
This baseline is a variant of our model, omitting the group-level temporal model (LSTM 2). In other words, we do the final classification based on the outputs of the temporal models for individual person action labels, but without an additional group-level LSTM.

\end{enumerate}

\subsection{Experiments on the Collective Activity Dataset}
The Collective Activity Dataset \cite{choi2009they} has been widely used for evaluating group activity recognition approaches in the computer vision literature \cite{amer2014hirf, DengZCLMRM15, amer2012cost}. This dataset consists of 44 videos, eight person-level pose labels (not used in our work), five person level action labels, and five group-level activities. A scene is assigned a group activity label based on the majority of what people are doing. We follow the train/test split provided by \cite{hajimirsadeghi2015visual}. In this section, we present our results on this dataset.

{\bf Model details:} In the Collective Activity Dataset, 9 timesteps and 3000 hidden nodes are used for the first LSTM layer and a softmax layer is deployed for the classification layer in this stage. The second network consists of a 3000-node fully connected layer followed by a 9-timestep 500-node LSTM layer which is passed to a softmax layer trained to recognize group activity labels.

  \begin{table}[ht]
  \begin{center}
  \begin{tabular}{|l|c|}
  \hline
  Method & Accuracy\\
  \hline
  \hline
  B1-Image Classification                  & 63.0 \\
  B2-Person Classification                 & 61.8 \\ 
  B3-Fine-tuned Person Classification      & 66.3 \\ \hline
  B4-Temporal Model with Image Features    & 64.2 \\
  B5-Temporal Model with Person Features   & 64.0 \\ \hline
  B6-Two-stage Model without LSTM 1   & 70.1 \\
  B7-Two-stage Model without LSTM 2   & 76.8 \\ \hline
  \bf Two-stage Hierarchical Model	  & \bf 81.5	\\ \hline
  \end{tabular}
  \end{center}
  \caption{Comparison of our method with baseline methods on the Collective Activity Dataset.}
  \label{tab:acc_cad}
  \end{table}
  
  \begin{table}[ht]
  \begin{center}
  \begin{tabular}{|l|c|}
  \hline
  Method & Accuracy\\
  \hline
  \hline
  Contextual Model \cite{LanWYRM12}         &   79.1 \\ \hline
  Deep Structured Model \cite{DengZCLMRM15} &   80.6 \\ \hline
  \bf Our Two-stage Hierarchical Model & 81.5 \\ \hline
  Cardinality kernel \cite{hajimirsadeghi2015visual} &	\bf 83.4 \\ \hline
  \end{tabular}
  \end{center}
  \caption{Comparison of our method with previously published works on the Collective Activity Dataset.}
  \label{acc_cad_comp}
  \end{table}
  
\begin{figure*}[ht]
\centering
\def\tabularxcolumn#1{m{#1}}
\begin{tabular}{cc}
\vspace{-0.2cm}
\subfloat[]{\includegraphics[width=4.5cm]{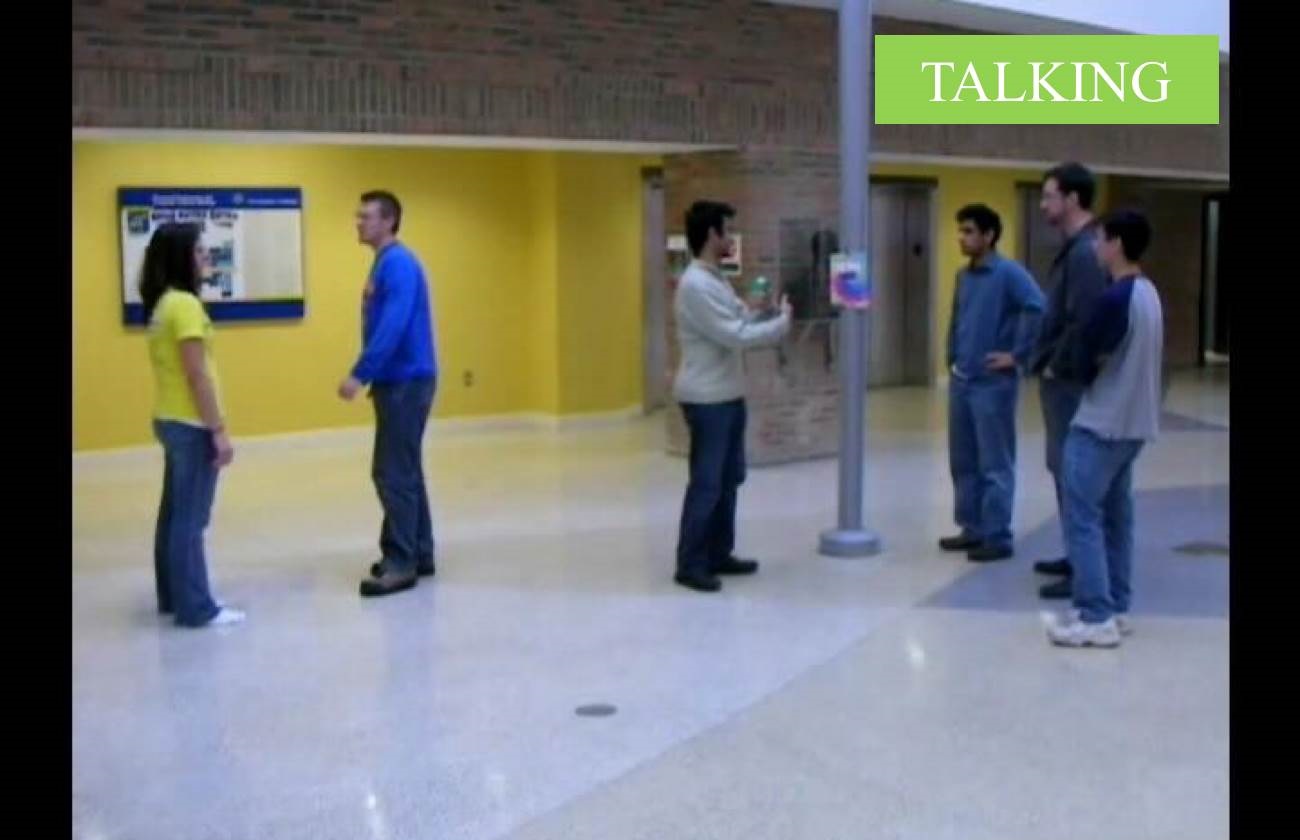}} 
   & \subfloat[]{\includegraphics[width=4.5cm]{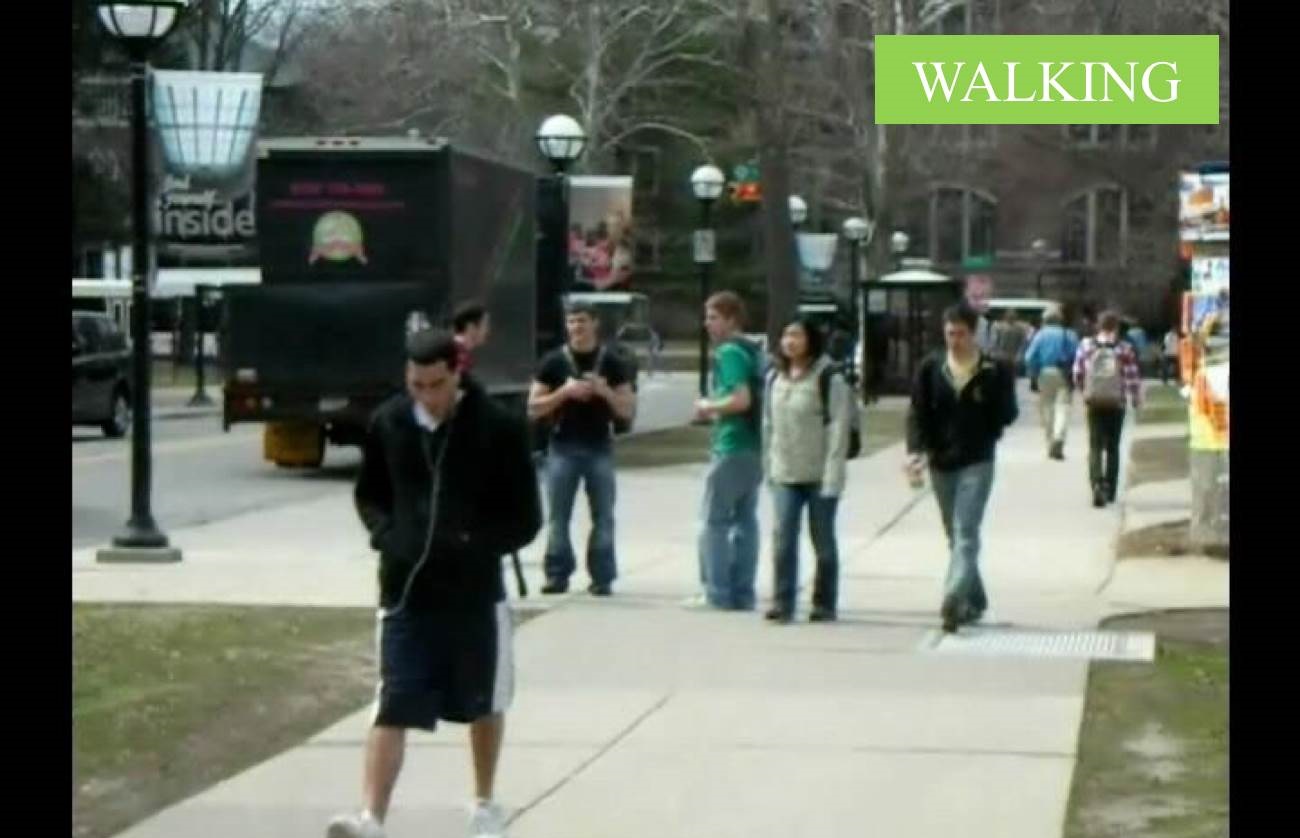}}\\ 
\vspace{-0.2cm}
\subfloat[]{\includegraphics[width=4.5cm]{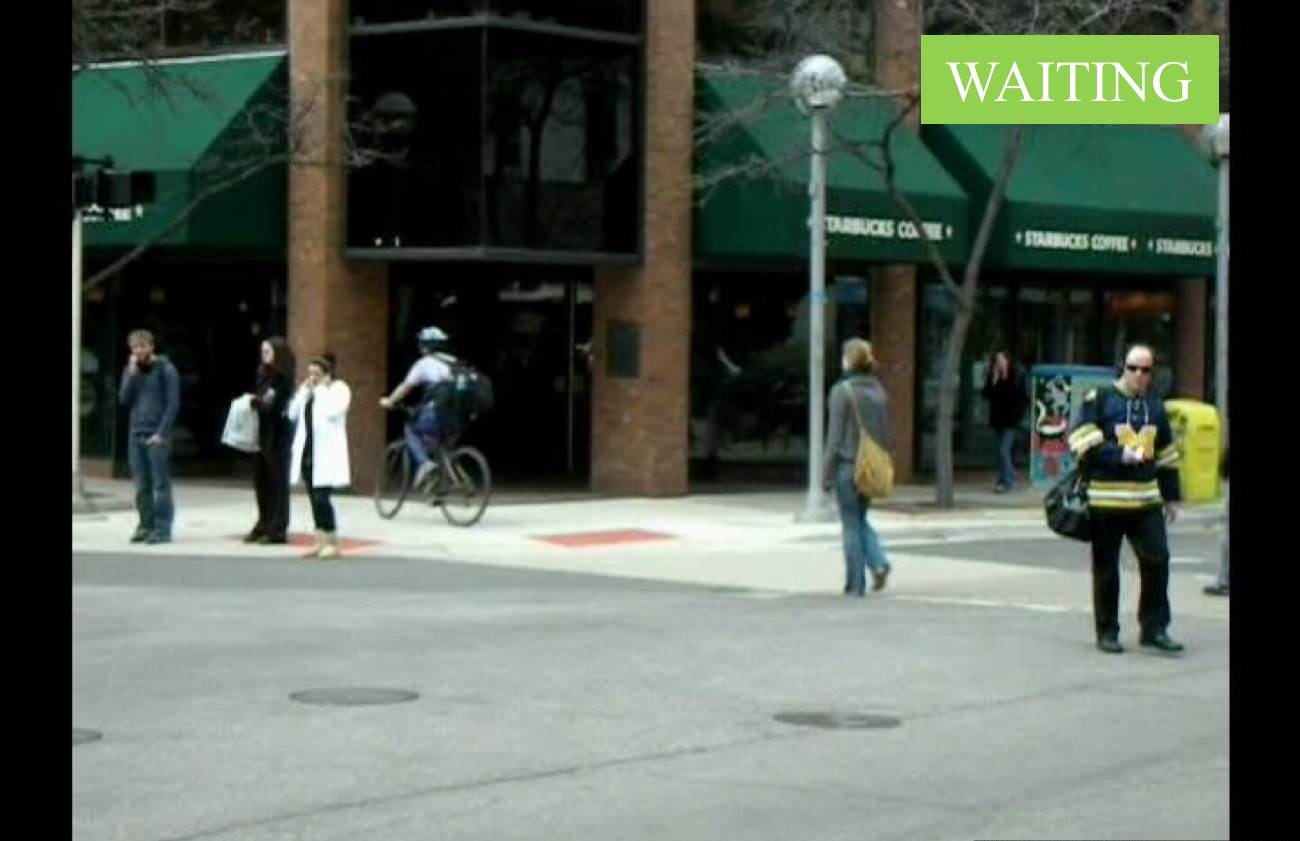}} 
   & \subfloat[]{\includegraphics[width=4.5cm]{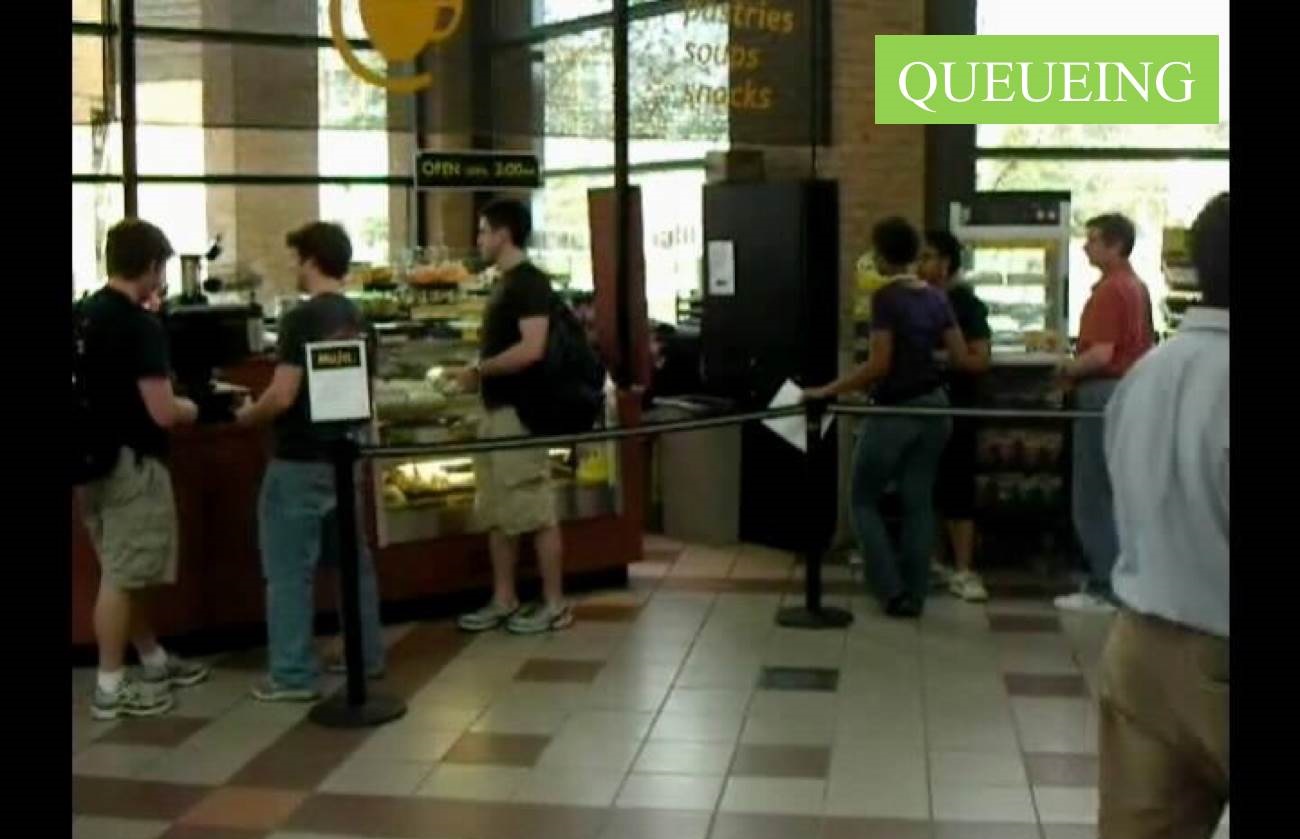}}\\
 \vspace{-0.2cm}
 \subfloat[]{\includegraphics[width=4.5cm]{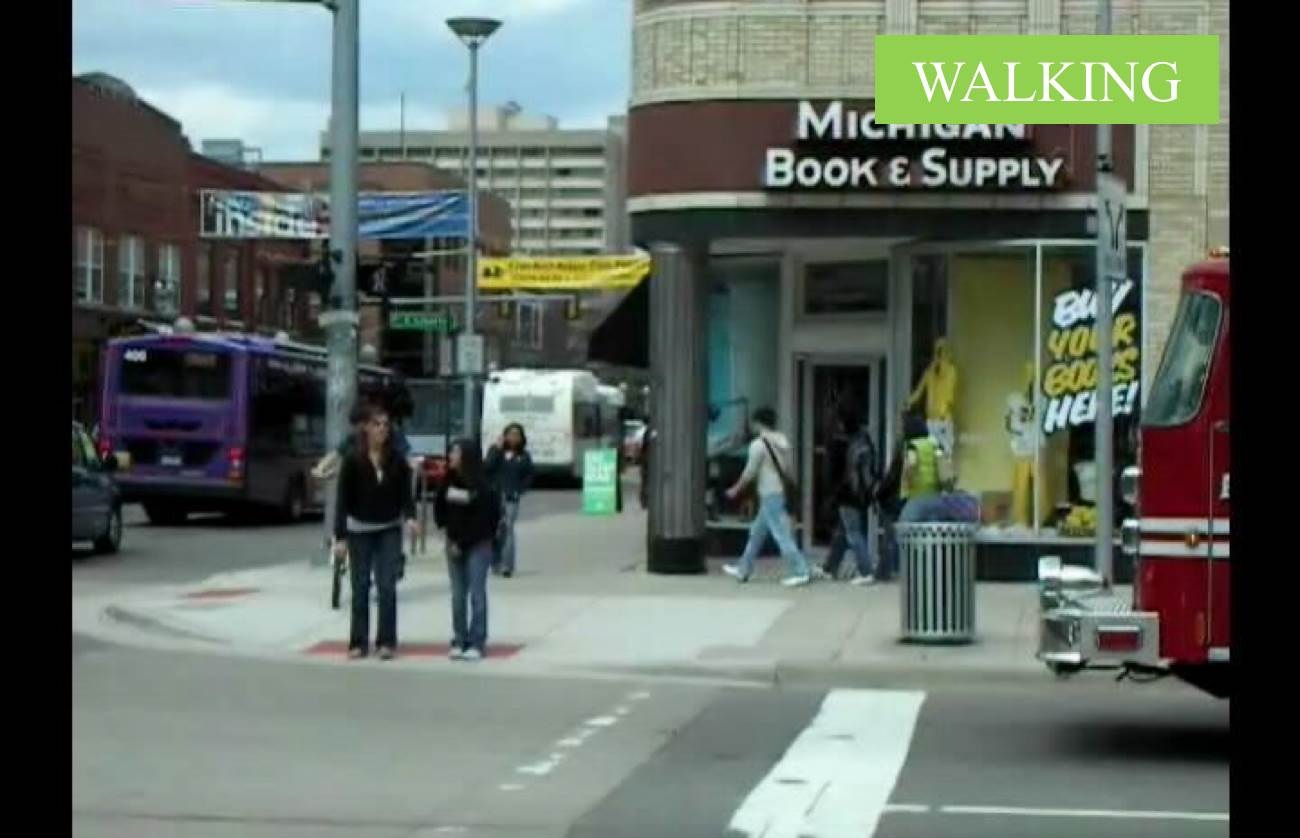}} 
    & \subfloat[]{\includegraphics[width=4.5cm]{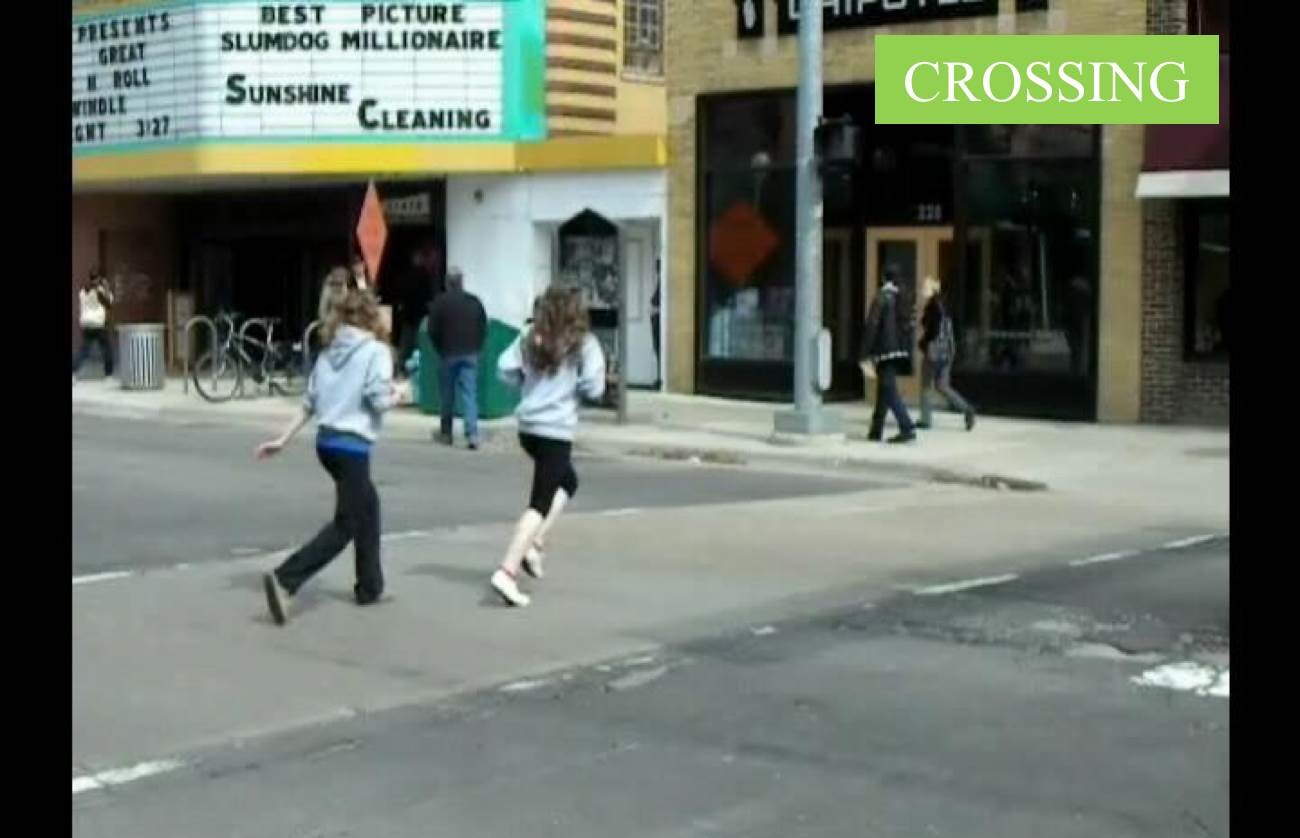}}\\
\vspace{-0.2cm}
\subfloat[]{\includegraphics[width=4.5cm]{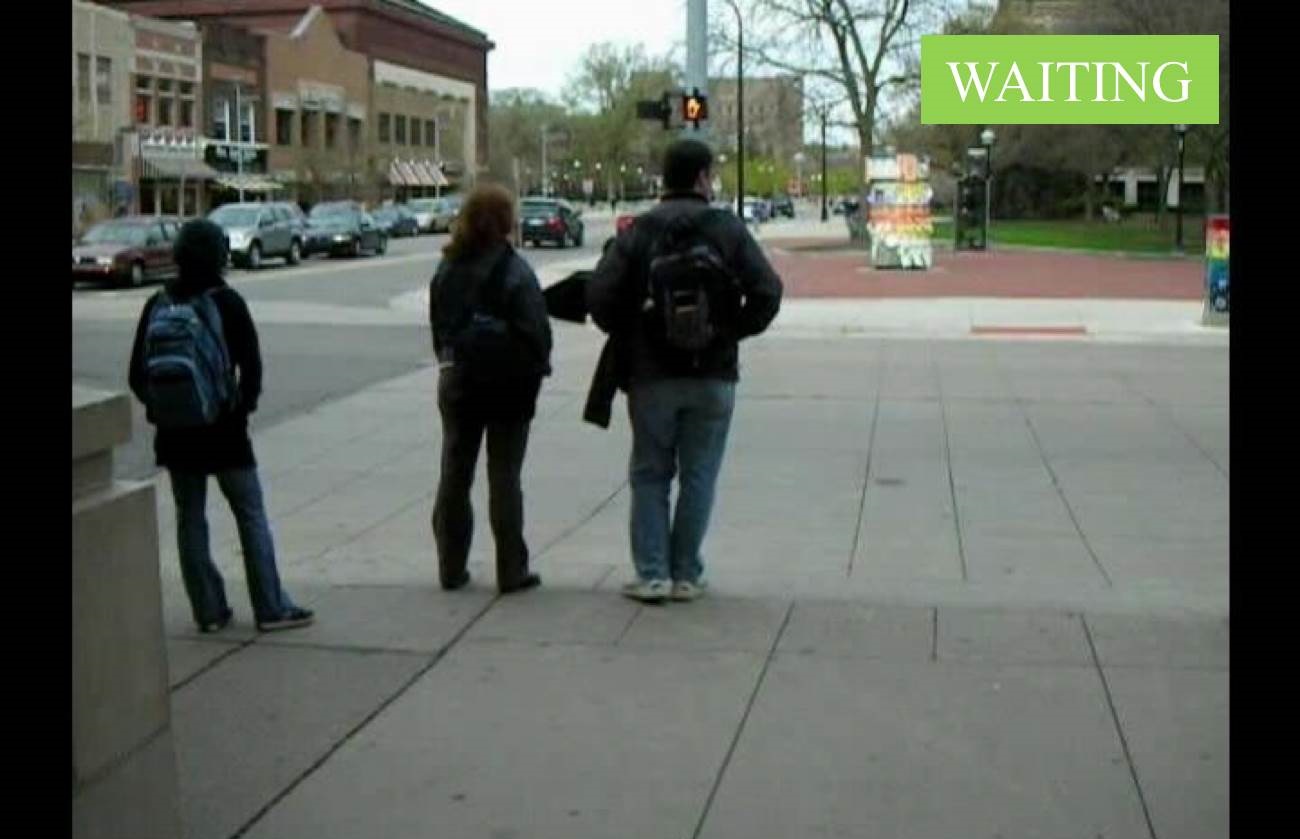}} 
   & \subfloat[]{\includegraphics[width=4.5cm]{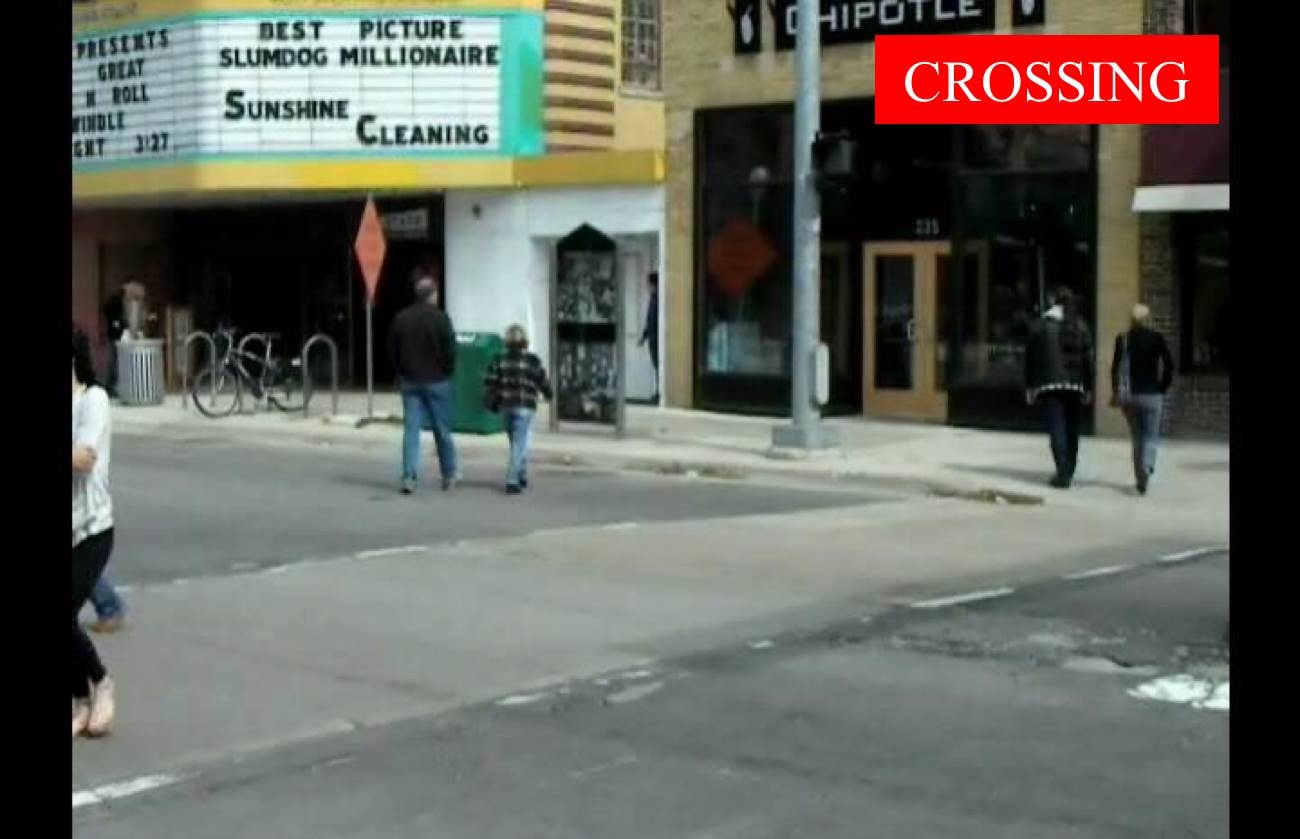}}\\
\vspace{-0.2cm}
\subfloat[]{\includegraphics[width=4.5cm]{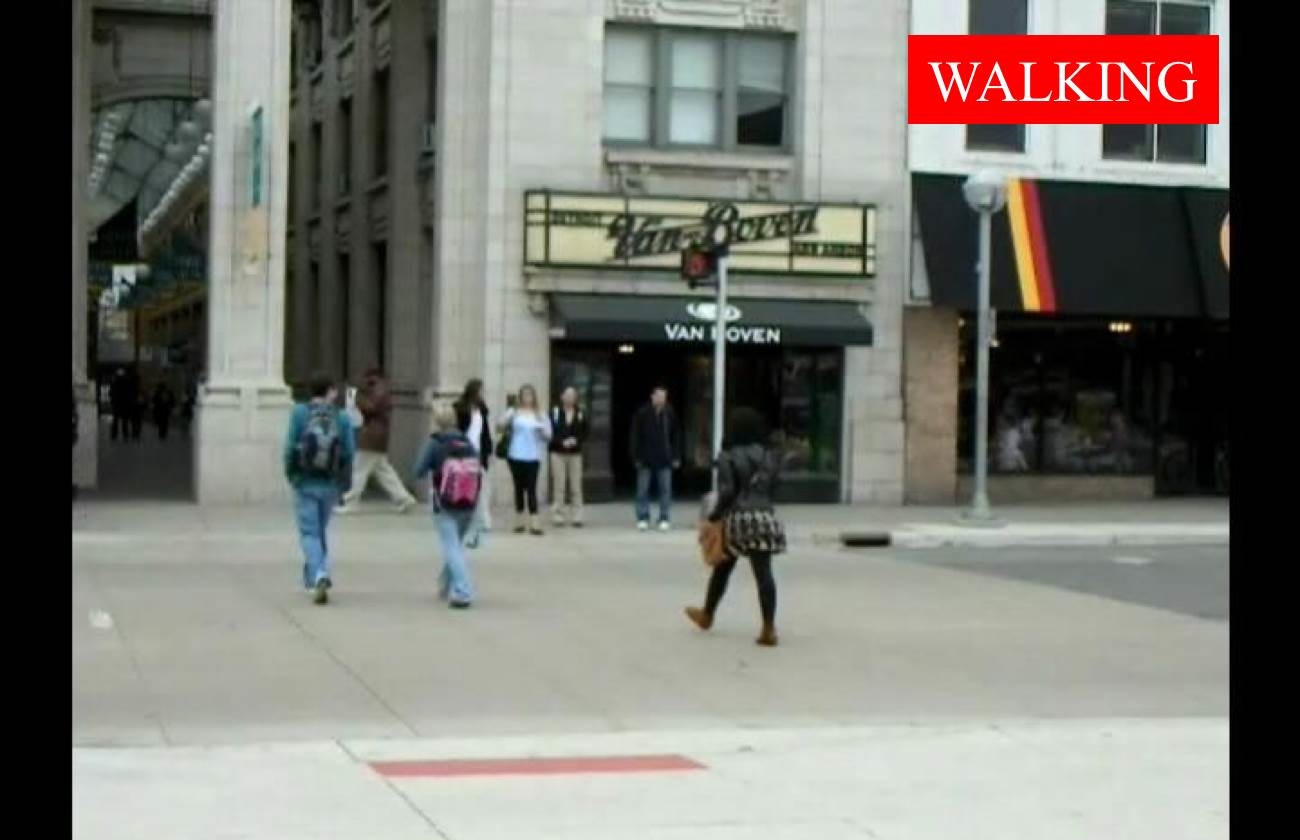}} 
   & \subfloat[]{\includegraphics[width=4.5cm]{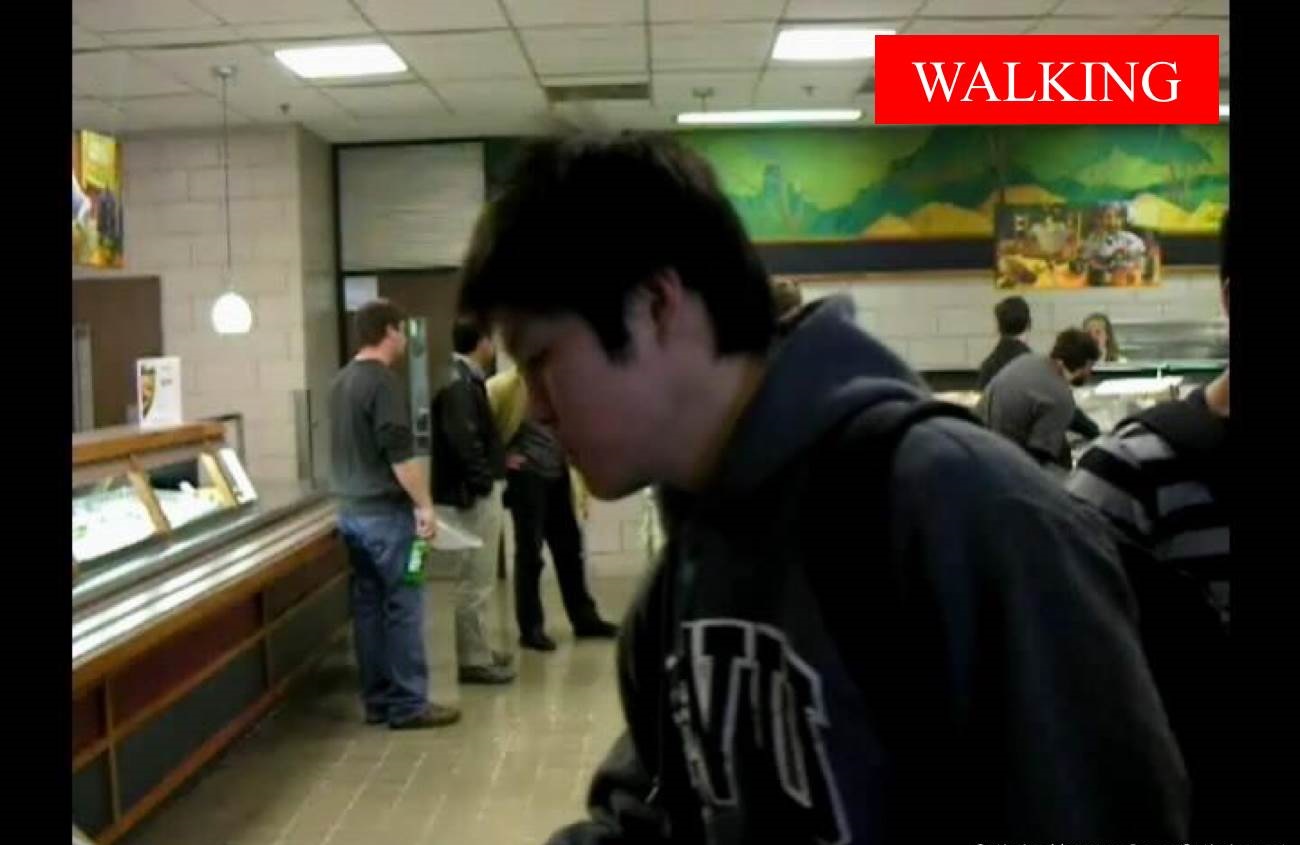}}\\
\end{tabular}

\caption{Visualizations of the generated scene labels from the Collective Activity Dataset using our model. Green denotes correct classifications, red denotes incorrect. The incorrect ones correspond to the confusion between different actions in ambiguous cases (h and j examples), or in the cases where there is an anomalous camera zoom.}\label{fig:vis_cad}
\end{figure*}

\begin{figure*}[t]
  \begin{center}
  \includegraphics[trim=0 0 0 0,clip, width=0.75\linewidth]{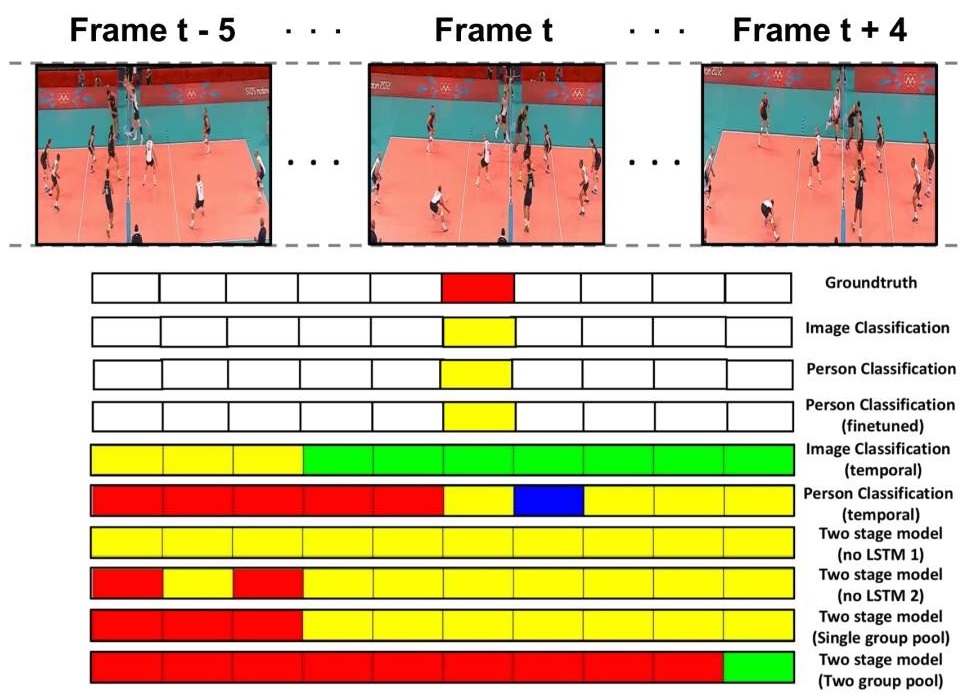}
  \end{center}
    \caption{Visualization of the generated labels by different baselines/models for a sample video extracted from the Volleyball Dataset. In this figure, yellow, red, blue and green colors denote the right spike, left pass, left spike, and left set group activities respectively.}
  \label{timeline}
  \end{figure*}

\begin{figure*}[t]
  \begin{center}
  \includegraphics[trim=0 0 0 0,clip, width=0.75\linewidth]{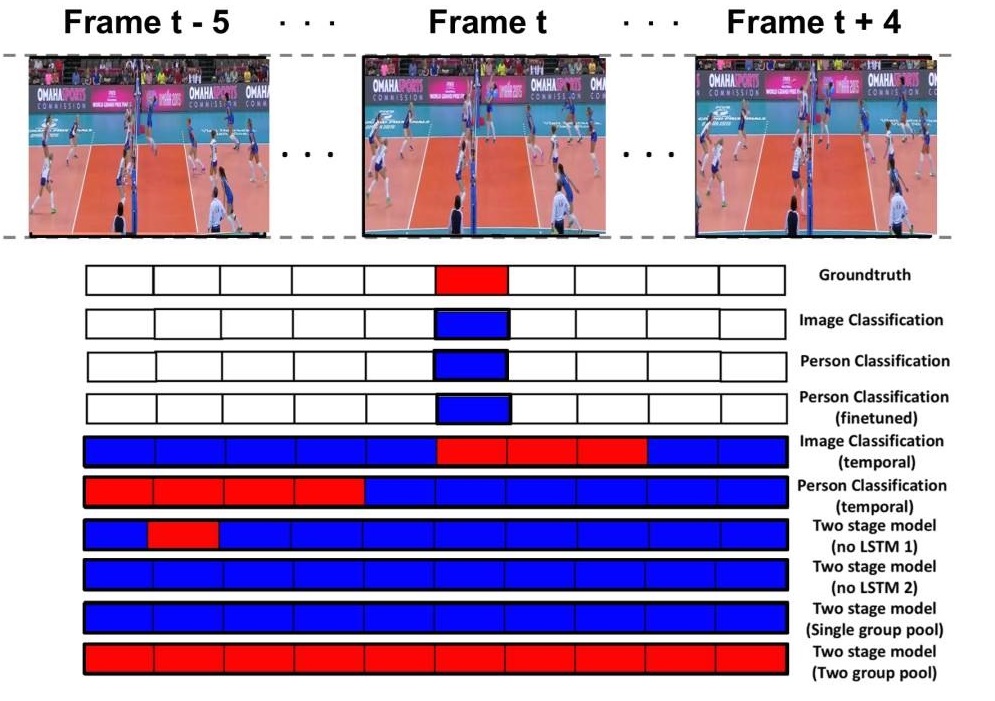}
  \end{center}
    \caption{Visualization of the generated labels by different baselines/models for another sample video extracted from the Volleyball Dataset. In this figure, red, blue colors denote the right spike and right set group activities respectively.}
  \label{timeline2}
  \end{figure*}
  
  {\bf Ablation studies:} In Table~\ref{tab:acc_cad}, the classification results of our proposed architecture is compared with the baselines. As shown in the table, our two-stage LSTM model significantly outperforms the baseline models.
  A comparison can be made between temporal and frame-based counterparts including B1 vs.\ B4, B2 vs.\ B5 and B3 vs.\ our two-stage model. We observe that adding temporal information using LSTMs improves the performance of these baselines. 
  
  {\bf Comparison to other methods:} Table~\ref{acc_cad_comp} compares our method with state of the art methods for group activity recognition. Fig.~\ref{fig:vis_cad} provides visualizations of example results. The performance of our two-stage model is comparable to the state of the art methods. Note that only Deng et al.~\cite{DengZCLMRM15} is a previously published deep learning model.  In contrast, the cardinality kernel approach~\cite{hajimirsadeghi2015visual} outperformed our model. It should be noted that this approach works on hand crafted features fed to a model highly optimized for a cardinality problem (i.e.\ counting the number of actions in the scene) which is exactly the way group activities are defined in this dataset.
  
  \subsubsection{Discussion}
 
  The confusion matrix obtained for the Collective Activity Dataset using our two-stage model is shown in Figure~\ref{fig:conf_mat_cad_mx_pool}. We observe that the model performs almost perfectly for the talking and queuing classes, and gets confused between crossing, waiting, and walking. Such behaviour is perhaps due to a lack of consideration of spatial relations between people in the group, which is shown to boost the performance of previous group activity recognition methods: e.g.\ crossing involves the walking action, but is confined in a path which people perform in orderly fashion. Therefore, our model that is designed only to learn the dynamic properties of group activities often gets confused with the walking action.
  
  It is clear that our two-stage model has improved performance with compared to baselines. The temporal information improves performance. Further, finding and describing the elements of a video (i.e.\ persons) provides benefits over utilizing frame level features.
  
  \begin{figure}
  \begin{center}
  \includegraphics[width=0.8\linewidth]{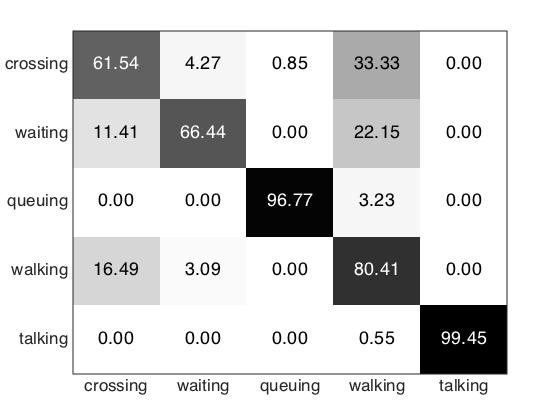}
    \caption{Confusion matrix for the Collective Activity Dataset obtained
    using our two-stage model.}
  \label{fig:conf_mat_cad_mx_pool}
  \end{center}
  \end{figure}

 \subsection{Experiments on the Volleyball Dataset}

In order to evaluate the performance of our model for team activity recognition on sport footage, we collected a new dataset using publicly available YouTube volleyball videos. We annotated 4830 frames that were handpicked from 55 videos with nine player action labels and eight team activity labels. We used frames from 2/3$^{rd}$ of the videos for training, and the remaining 1/3$^{rd}$ for testing. The list of action and activity labels and related statistics are tabulated in Tables \ref{scene_voll} and \ref{action_tab}.

From the tables, we observe that the group activity labels are relatively more balanced compared to the player action labels.  This follows from the fact that we often have people present in static actions like standing compared to dynamic actions (setting, spiking, etc.). Therefore, our dataset presents a challenging team activity recognition task, where we have interesting actions that can directly determine the group activity occur rarely in our dataset. The dataset will be made publicly available to facilitate future comparisons \footnote{\url{https://github.com/mostafa-saad/deep-activity-rec}}.

{\bf Model details:} The model hyperparameters for the Volleyball Dataset include 5 timesteps and 3000 hidden nodes for the first LSTM layer. The second network uses 10 timesteps and 2000 hidden nodes for the second LSTM layer.

We further experiment with a set of different player sub-grouping approaches for pooling.  To find the sub-groups, we follow a simple strategy. First, we order players based on their top-left bounding box point (x-axis first). To split players to two groups (e.g. left/right teams), we consider the first half of players as group one. Similarly, to split to four groups, we consider the first quarter of players as group one, second quarter as group two and so on. If players cannot be divided evenly (missing players), the last sub-groups will have fewer players.

 \begin{table}[ht]
 \begin{center}
  \begin{tabular}{|l|c|}
  \hline
  Group Activity Class & \# Instances\\
  \hline
  \hline
   Right set			&	644	\\ \hline
   Right spike			&	623	\\ \hline
   Right pass           &   801 \\ \hline
   Right winpoint       &   295 \\ \hline
   Left winpoint        &   367 \\ \hline
   Left pass	    	&	826	\\ \hline
   Left spike 	    	&	642	\\ \hline
   Left set	            &	633 \\	\hline
  \end{tabular}
  \caption{Statistics of the group activity labels in the Volleyball Dataset.}
  \label{scene_voll}
  \end{center}
\end{table}

\begin{table}
\begin{center}
    \begin{tabular}{|l|c|}
  \hline
  Action Class & \# Instances\\
  \hline
  \hline
   Waiting          &   3601 \\ \hline
   Setting		    &	1332 \\ \hline
   Digging          &   2333 \\ \hline
   Falling          &   1241 \\ \hline
   Spiking          &   1216 \\ \hline
   Blocking         &   2458 \\ \hline
   Jumping          &   341 \\ \hline
   Moving           &   5121 \\ \hline
   Standing         &   38696 \\ \hline
    
  \end{tabular}
   \caption{Statistics of the action labels in the Volleyball Dataset.}
  \label{action_tab}
  \end{center}
  \end{table}

  %In our experiment, we wished our first stage LSTM to learn representations pertaining to the important actions. For this purpose, we used six labels for the first stage instead of nine, by ignoring unimportant actions that include standing, moving and jumping. In the person level feature extraction stage, we allowed our model to overfit on persons who were performing these actions. This step was essential for us to focus on key actions for recognizing group activities.
  
  {\bf Ablation studies:} In Table \ref{tab:acc_vol_2_groups}, the classification performance of our proposed model is compared against the baselines. Similar to the performance in the Collective Activity Dataset, our two-stage LSTM model outperforms the baseline models. 
  
  %However, compared to the baselines, the performance gain using our model is more modest. This is likely because we can infer group activity in volleyball by using just a few frames. 
  %Therefore, in the Volleyball Dataset, our baseline B1 is closer to the actual model's performance, compared to the Collective Activity Dataset. 
  
  Moreover, explicitly modeling people is necessary for obtaining better performance in this dataset, since the background is rapidly changing due to a fast moving camera, and therefore it corrupts the temporal dynamics of the foreground. This could be verified from the performance of our baseline model B4, which is a temporal model that does not consider people explicitly, showing inferior performance compared to the baseline B1, which is a non-temporal image classification style model. On the other hand, baseline model B5, which is a temporal model that explicitly considers people, performs comparably to the image classification baseline, in spite of the problems that arise due to tracking and motion artifacts. 
  
  \begin{table}[ht]
  \begin{center}
  \begin{tabular}{|l|c|}
  \hline
  Method & Accuracy\\
  \hline
  \hline
  B1-Image Classification                      & 66.7 \\
  B2-Person Classification                     & 64.6 \\ 
  B3-Fine-tuned Person Classification          & 68.1 \\ \hline
  B4-Temporal Model with Image Features        & 63.1 \\
  B5-Temporal Model with Person Features       & 67.6 \\ \hline
  B6-Two-stage Model without LSTM 1            & 74.7 \\
  B7-Two-stage Model without LSTM 2            & 80.2 \\ \hline
  \bf Our Two-stage Hierarchical Model     & \bf 81.9 \\ \hline
  \end{tabular}
  \end{center}
  \caption{Comparison of the team activity recognition performance of baselines against our model evaluated on the Volleyball Dataset. Experiments are using 2 group styles with max pool strategy.}
  \label{tab:acc_vol_2_groups}
  \end{table}

In both datasets, an observation from the tables is that while both LSTMs contribute to overall classification performance, having the first layer LSTM (B7 baseline) is relatively more critical to the performance of the system, compared to the second layer LSTM (B6 baseline). 

To further investigate players sub-grouping, in Table \ref{tab:acc_vol_groups_1_2_4}, we run experiments over 4 sub-groups: left-team-back players, left-team-front players, right-team-back players and front-team-bottom players. 

It seems from the results that more brute force sub-grouping doesn't improve the performance of the system for this dataset. It shows that extracting additional information by segregating players on basis of their position renders information from static/insignificant players results in more confusion, and perhaps leading to degradation in performance. Therefore, from this experiment, it is evident all types of explicit spatio-temporal relation modelling does not lead to an improvement in performance.

  \begin{table}[ht]
  \begin{center}
  \begin{tabular}{|l|c|}
  \hline
  Method & Accuracy\\
  \hline
  \hline
  Our Model - 1 group - max pool            & 70.3 \\
  Our Model - 1 group - avg pool            & 68.5 \\  \hline
  Our Model - 2 groups - max pool       & \bf 81.9 \\
  Our Model - 2 groups - avg pool           & 80.7 \\  \hline
  Our Model - 4 groups - max pool           & 81.5 \\
  Our Model - 4 groups - avg pool           & 79.6 \\  \hline
  \end{tabular}
  \end{center}
  \caption{Comparison of the team activity recognition of our model using 2 sub-groups vs.\ 4 sub-groups with both average and max pooling.}
  \label{tab:acc_vol_groups_1_2_4}
  \end{table}

To evaluate the effect of number of LSTM nodes of the model's two network, we conducted set of experiments outlined in Table~\ref{tab:acc_vol_hidden1_2}.  Similarly, we evaluate the effect of the number of timesteps of the model's two network, we conducted set of experiments outlined in Table~\ref{tab:acc_vol_timesteps1_2}.

  \begin{table}[ht]
  \begin{center}
  \begin{tabular}{|l|c|c|c|}
  \hline
  Method & No. Person & No. Scene  & Accuracy\\
         & LSTM Nodes & LSTM Nodes &         \\
  \hline
  \hline
  Our Model & 1000 & 1000           & 79.4 \\
  Our Model & 2000 & 1000           & 80.3 \\
  Our Model & 3000 & 1000           & 81.2 \\ \hline
  Our Model & 3000 & 2000       & \bf 81.9 \\
  Our Model & 3000 & 3000          &  81.2 \\ \hline
  \end{tabular}
  \end{center}
  \caption{Comparison of the team activity recognition of our model using 2 groups style over different numbers of LSTM nodes in the second, group-level LSTM layer.}
  \label{tab:acc_vol_hidden1_2}
  \end{table}

  \begin{table}[ht]
  \begin{center}
  \begin{tabular}{|l|c|c|c|}
  \hline
  Method & No. Person & No. Scene  & Accuracy\\
         & timesteps  & timesteps  &         \\
  \hline
  \hline
  Our Model & 5 & 10      & \bf 81.9       \\ \hline
  Our Model & 5 & 20          & 81.7  \\ \hline
  Our Model & 10 & 10         & 81.7  \\ \hline
  Our Model & 10 & 20         & 81.3  \\ \hline
  \end{tabular}
  \end{center}
  \caption{Comparison of the team activity recognition of our model using 2 groups style over different number of timesteps in the model 2 networks}
  \label{tab:acc_vol_timesteps1_2}
  \end{table}
  
{\bf Comparison to other methods:} In Table~\ref{tab:acc_vol_idtf}, we compare our model to the improved dense trajectory approach~\cite{Wang2013}.  Dense trajectories is a hand-crafted approach that competes strongly versus deep learning features. In addition, we also created two variations of~\cite{Wang2013}, where the considered trajectories are only the ones inside the players' bounding boxes, in other words, ignoring background trajectories. The variations emulate our model with one group and 2 groups style. That is, the first variation represents the players from the whole team, while the second represents each team and then concatenates the two representations to get the whole scene representation.

  \begin{table}[ht]
  \begin{center}
  \begin{tabular}{|l|c|}
  \hline
  Method & Accuracy\\
  \hline
  \hline
  Our Model using 2 groups style                       & \bf 81.9 \\  \hline
  IDTF~\cite{Wang2013} - all trajectories                  & 73.4 \\
  IDTF~\cite{Wang2013} - 1 group-box trajectories          & 71.7 \\
  IDTF~\cite{Wang2013} - 2 groups-box trajectories         & 78.7 \\  \hline
  \end{tabular}
  \end{center}
  \caption{Comparison of the team activity recognition of our model against improved dense trajectory approach approach.}
  \label{tab:acc_vol_idtf}
  \end{table}

The traditional dense trajectories approach and its one group style show close performance, but the 2-groups trajectories variation yields higher performance. Probably, this is due the reduction of confusions between left and right teams' activities. However, our model outperforms these strong  dense trajectory-based baseline methods.

  %In the Volleyball Dataset, we consider 2 baselines versus our structured model. In the first baseline experiment, we learn pure scene classification: categorize a frame to one of the group activities with no temporal information considered. To train such network, we fine-tuned AlexNet with training examples of (frame, activity label) pairs. \\

  %In the second baseline, we do also scene classification but though a temporal window over K tracklets. Specifcally, we detect K players in the target testing frame and track them through window of W frames centered at that frame (hence capture information before and after the event). A W-reshaped frames for the K bounding boxes are built as in Figure~\ref{fig_detect_reshape_baseline2}. To train a spatio-temporal model for that, we feed the reshaped frames to a network composed of AlexNet followed by LSTM layer. 

%\begin{figure}
%    \centering
%    \subfloat[Detections]{{\includegraphics[height=2.5cm]{images/00026591.jpg} }}
%    \hspace{0.7pt}   
%    \subfloat[Reshaped Image]{{\includegraphics[height=2.5cm,width=3cm]{images/00026591_reshape.jpg} }}
%    \caption{Baseline 2: players reshaping. In the left image, we detect the K people and then reshape them as in the right image. As classifier is not accuracte, K is slightly bigger than the actual number of players to avoid missing them.}
%    \label{fig_detect_reshape_baseline2}
%\end{figure}

\subsubsection{Discussion}

Figure~\ref{fig:conf_mat_voll_1g_mx_pool} shows the confusion matrix obtained for the Volleyball Dataset using our two-stage model by grouping all players (no sub-groups) in one representation using max pooling operation, similar to~\cite{msibrahi16deepactivity}. From the confusion matrix, we observe that our model generates accurate high level action labels. Nevertheless, our model has some confusion between {\em left winpoint} and {\em right winpoint} activities. On the contrary to ~\cite{msibrahi16deepactivity}, the confusion between set and pass activities is resolved, probably due to using more data.

\begin{figure}
  \includegraphics[width=1.0\linewidth]{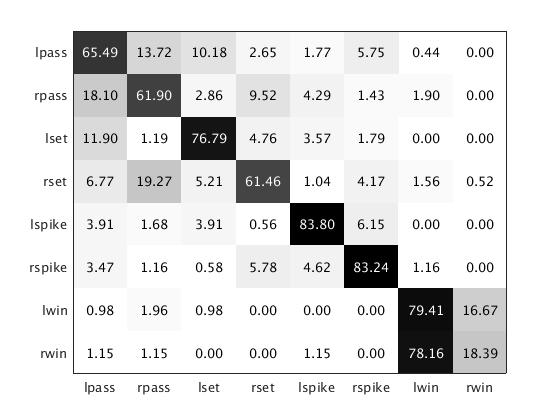}
    \caption{Confusion matrix for the Volleyball Dataset obtained
    using our two-stage hierarchical model, using 1 group style for all players.}
  \label{fig:conf_mat_voll_1g_mx_pool}
  \end{figure}

Figures~\ref{fig:conf_mat_voll_2g_mx_pool} shows the confusion matrix obtained for the Volleyball Dataset using our two-stage model, but by sub-grouping left team and right team first. From the confusion matrix, we observe that our model generates more accurate high level action labels than using no groups. In addition, the confusion between left winpoint and right winpoint activities is reduced.

\begin{figure}
  \includegraphics[width=1.0\linewidth]{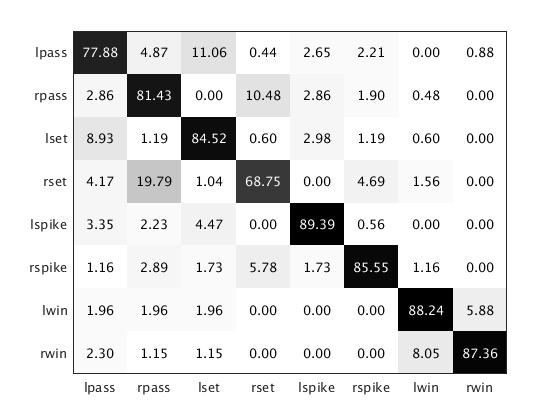}
    \caption{Confusion matrix for the Volleyball Dataset obtained
    using our two-stage hierarchical model, using 2 groups style.}
  \label{fig:conf_mat_voll_2g_mx_pool}
  \end{figure}

In Figure~\ref{fig:vis_voll}, we show the visualizations of our detected activities with different failure and success scenarios.
  
\begin{figure*}
\centering
\def\tabularxcolumn#1{m{#1}}
\begin{tabular}{cc}
\vspace{-0.2cm}
\subfloat[]{\includegraphics[width=6.5cm]{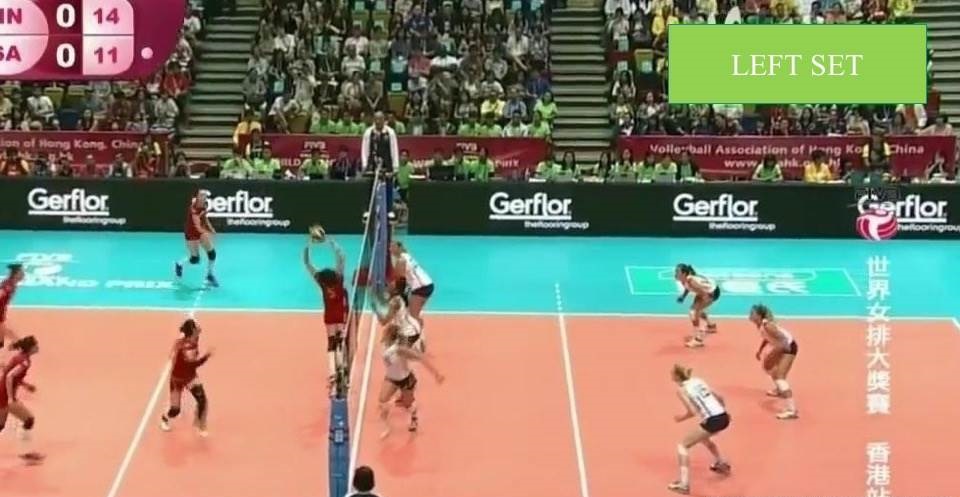}} 
   & \subfloat[]{\includegraphics[width=6.5cm]{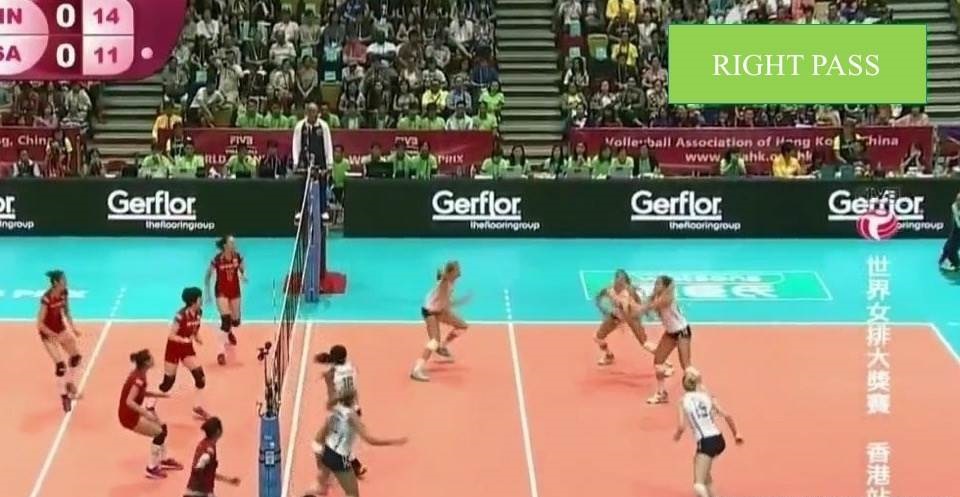}}\\ 
\vspace{-0.2cm}
\subfloat[]{\includegraphics[width=6.5cm]{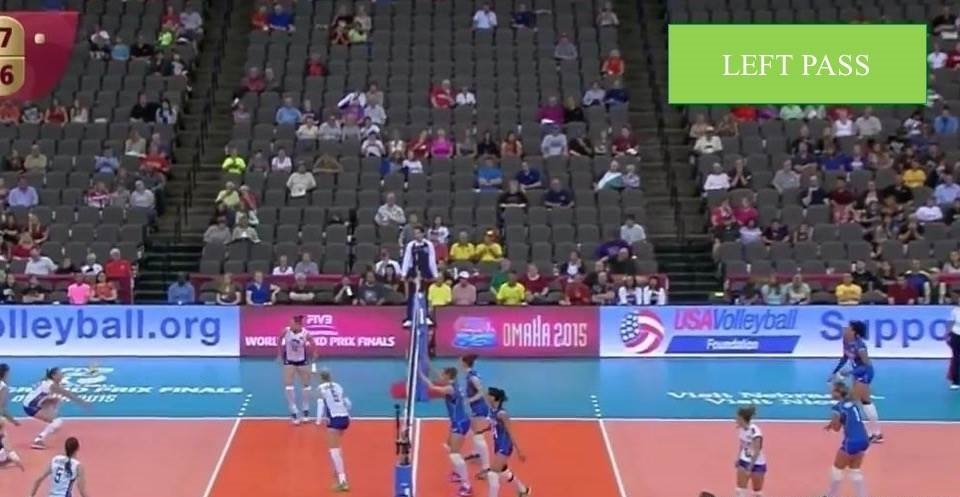}} 
   & \subfloat[]{\includegraphics[width=6.5cm]{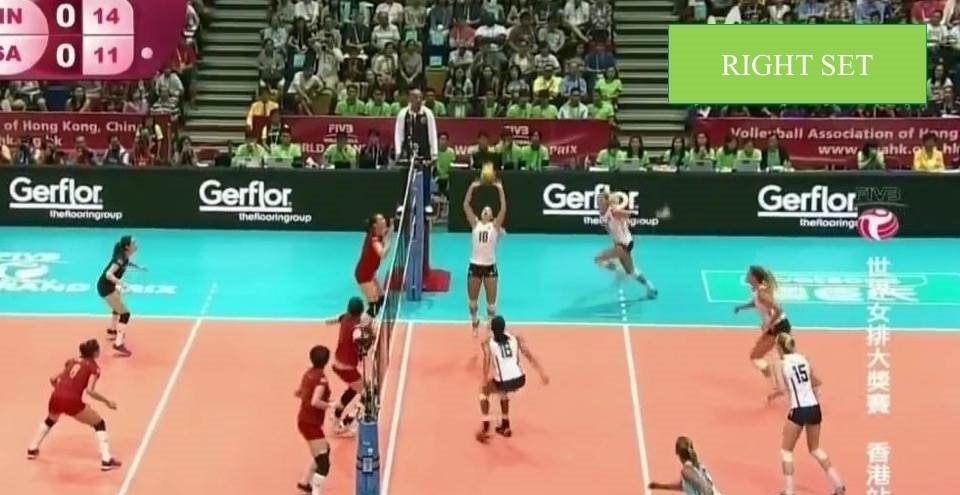}}\\
 \vspace{-0.2cm}
 \subfloat[]{\includegraphics[width=6.5cm]{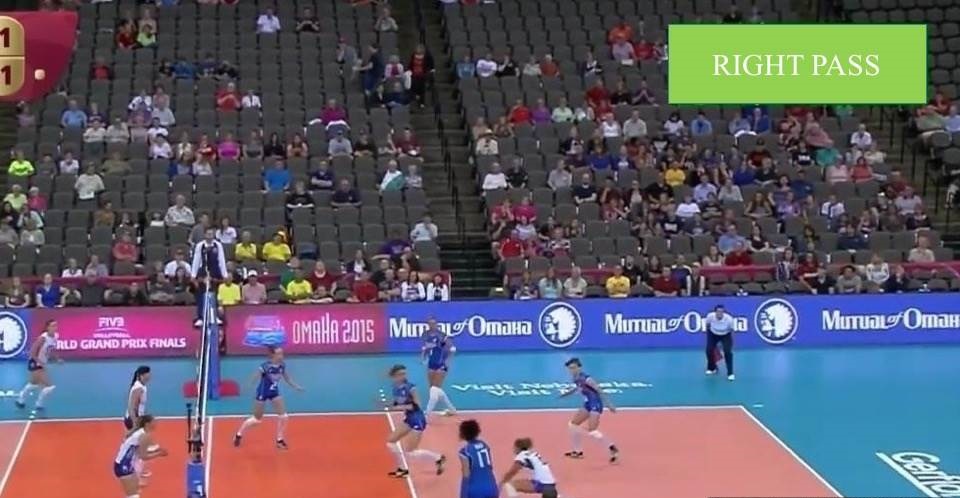}} 
    & \subfloat[]{\includegraphics[width=6.5cm]{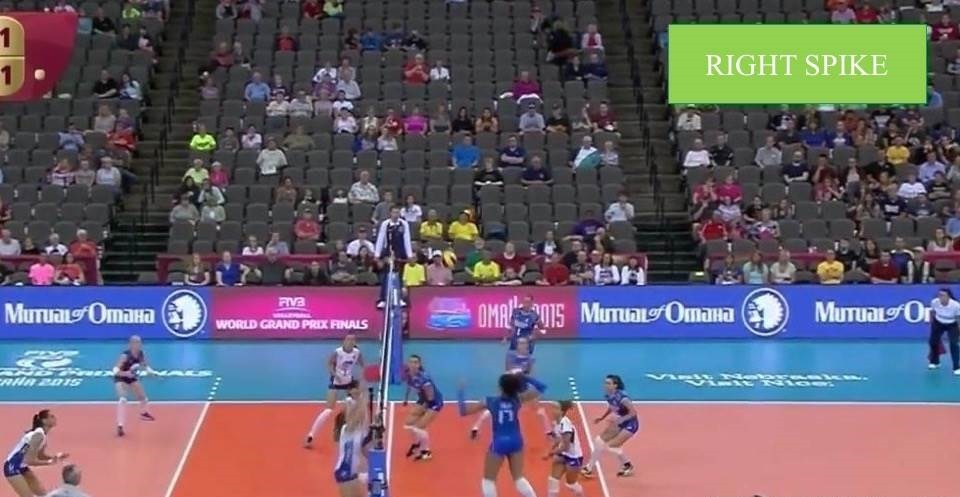}}\\
\vspace{-0.2cm}
\subfloat[]{\includegraphics[width=6.5cm]{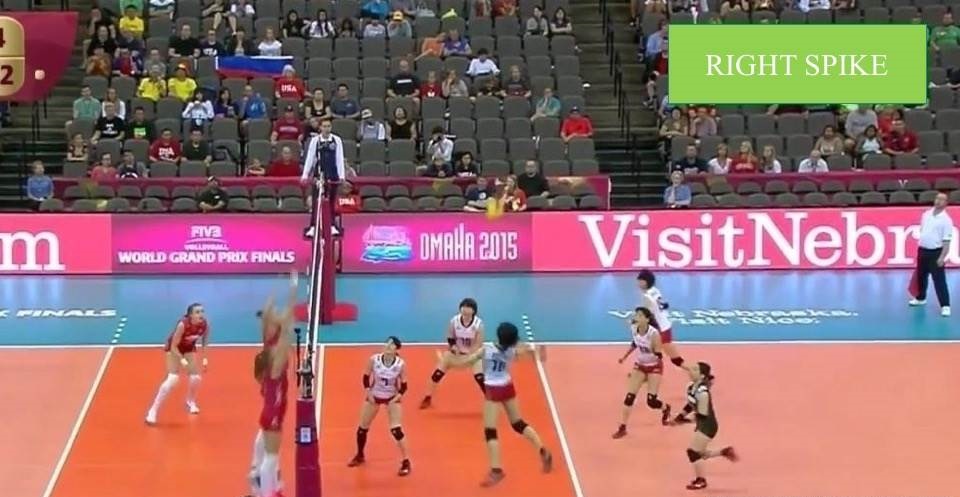}} 
   & \subfloat[]{\includegraphics[width=6.5cm]{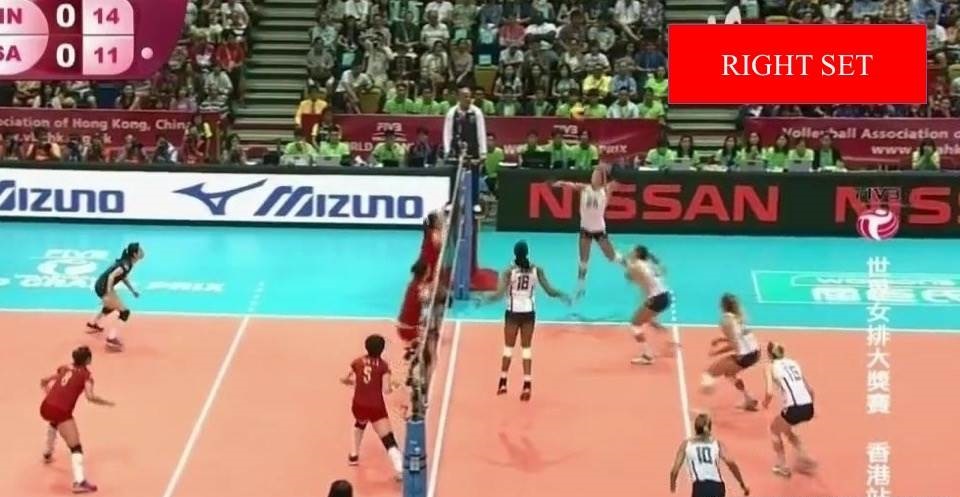}}\\
\vspace{-0.2cm}
\subfloat[]{\includegraphics[width=6.5cm]{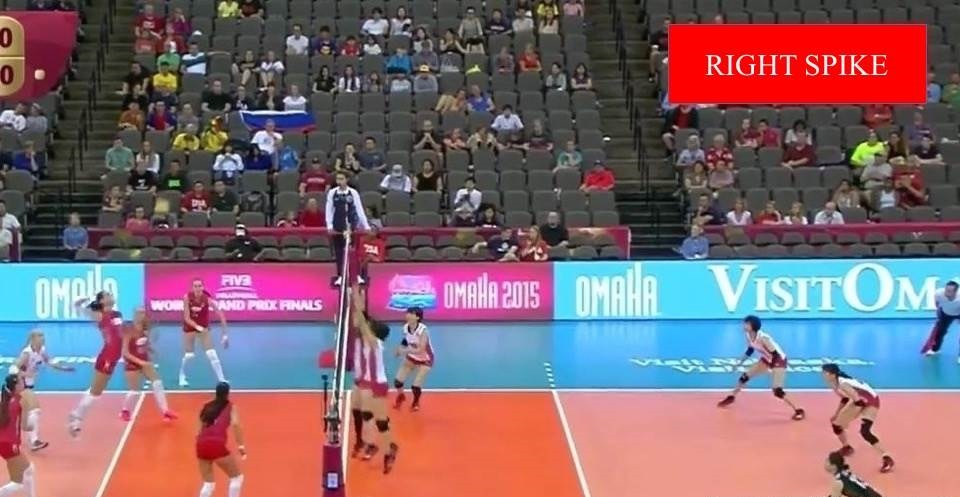}} 
   & \subfloat[]{\includegraphics[width=6.5cm]{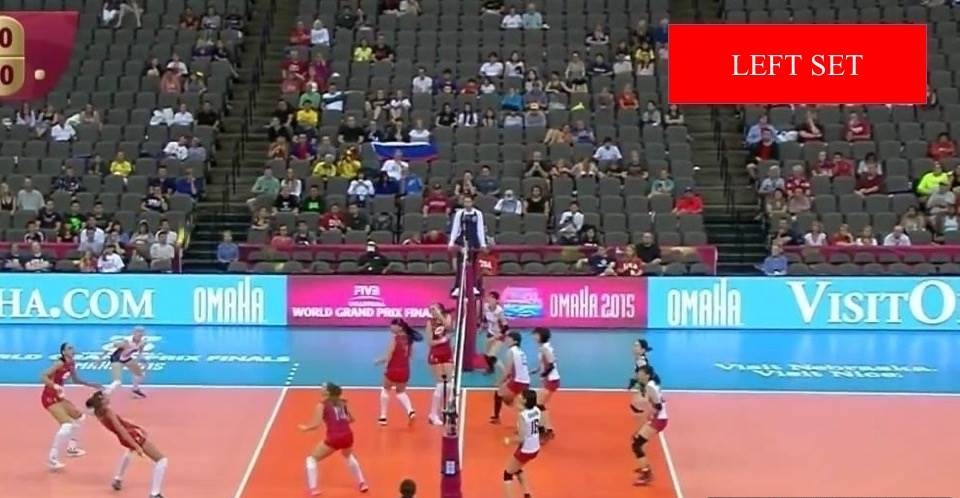}}\\
\end{tabular}

\caption{Visualizations of the generated scene labels from the Volleyball Dataset using our model.  Green denotes correct classifications, red denotes incorrect. The incorrect ones correspond to the confusion between different actions in ambiguous cases (h and j examples), or in the left and right distinction (i example).}\label{fig:vis_voll}
\end{figure*}
  
  %problem with identifying the side that is performing that action. {\color{red} This is evident from high values observed for the wrong side as shown in the confusion matrix.}

%  \begin{figure}[t]
%  \begin{center}
%  \includegraphics[width=0.8\linewidth]{images/00026591.jpg}
%  \end{center}
%    \caption{{\bf Visualization of our results on the Volleyball Dataset.} .... show some images, boxes in it, show the LSTM1 judging of actions and overall activity.}
%  \label{fig_compare_sucess_fail}
%  \end{figure}

%-------------------------------------------------------------------------
\section{Conclusion}
\label{sec:conclusion}
In this paper, we presented a novel deep structured architecture to deal with the group activity recognition problem. Through a two-stage process, we learn a temporal representation of person-level actions and combine the representation of individual people to recognize the group activity. We  created a new Volleyball Dataset to train and test our model, and also
evaluated our model on the Collective Activity Dataset. Results show that our architecture can improve upon baseline methods lacking hierarchical consideration of individual and group activities using deep learning.

%In the future work, we would like to try complex fusion styles and compare their different performance. e.g. We could improve the performance in volleyball dataset by a model that is explicitly designed to identify the side that performs the action. In addition, we plan to gather a bigger volleyball dataset and evaluate our models against it.

\section*{Acknowledgements}

This work was supported by grants from NSERC and Disney Research.

\appendices

\bibliographystyle{IEEEtran}
\bibliography{egbib.bib}

\begin{thebibliography}{10}
\providecommand{\url}[1]{#1}
\csname url@samestyle\endcsname
\providecommand{\newblock}{\relax}
\providecommand{\bibinfo}[2]{#2}
\providecommand{\BIBentrySTDinterwordspacing}{\spaceskip=0pt\relax}
\providecommand{\BIBentryALTinterwordstretchfactor}{4}
\providecommand{\BIBentryALTinterwordspacing}{\spaceskip=\fontdimen2\font plus
\BIBentryALTinterwordstretchfactor\fontdimen3\font minus
  \fontdimen4\font\relax}
\providecommand{\BIBforeignlanguage}[2]{{%
\expandafter\ifx\csname l@#1\endcsname\relax
\typeout{** WARNING: IEEEtran.bst: No hyphenation pattern has been}%
\typeout{** loaded for the language `#1'. Using the pattern for}%
\typeout{** the default language instead.}%
\else
\language=\csname l@#1\endcsname
\fi
#2}}
\providecommand{\BIBdecl}{\relax}
\BIBdecl

\bibitem{choi2009they}
W.~Choi, K.~Shahid, and S.~Savarese, ``What are they doing?: Collective
  activity classification using spatio-temporal relationship among people,'' in
  \emph{IEEE International Conference on Computer Vision Workshops (ICCV
  Workshops)}.\hskip 1em plus 0.5em minus 0.4em\relax IEEE, 2009, pp.
  1282--1289.

\bibitem{lan2012social}
T.~Lan, L.~Sigal, and G.~Mori, ``Social roles in hierarchical models for human
  activity recognition,'' in \emph{IEEE Conference on Computer Vision and
  Pattern Recognition (CVPR)}.\hskip 1em plus 0.5em minus 0.4em\relax IEEE,
  2012, pp. 1354--1361.

\bibitem{ramanathan2013social}
V.~Ramanathan, B.~Yao, and L.~Fei-Fei, ``Social role discovery in human
  events,'' in \emph{IEEE Conference on Computer Vision and Pattern Recognition
  (CVPR)}.\hskip 1em plus 0.5em minus 0.4em\relax IEEE, 2013, pp. 2475--2482.

\bibitem{amer2014hirf}
M.~R. Amer, P.~Lei, and S.~Todorovic, ``Hirf: Hierarchical random field for
  collective activity recognition in videos,'' in \emph{European Conference on
  Computer Vision (ECCV)}.\hskip 1em plus 0.5em minus 0.4em\relax Springer,
  2014, pp. 572--585.

\bibitem{wang2011action}
H.~Wang, A.~Kl{\"a}ser, C.~Schmid, and C.-L. Liu, ``Action recognition by dense
  trajectories,'' in \emph{IEEE Conference on Computer Vision and Pattern
  Recognition (CVPR)}.\hskip 1em plus 0.5em minus 0.4em\relax IEEE, 2011, pp.
  3169--3176.

\bibitem{schuldt2004recognizing}
C.~Sch{\"u}ldt, I.~Laptev, and B.~Caputo, ``Recognizing human actions: a local
  svm approach,'' in \emph{International Conference on Pattern Recognition,
  ICPR}, vol.~3.\hskip 1em plus 0.5em minus 0.4em\relax IEEE, 2004, pp. 32--36.

\bibitem{simonyan2014two}
K.~Simonyan and A.~Zisserman, ``Two-stream convolutional networks for action
  recognition in videos,'' in \emph{Neural Information Processing Systems
  (NIPS)}, 2014, pp. 568--576.

\bibitem{karpathy2014large}
A.~Karpathy, G.~Toderici, S.~Shetty, T.~Leung, R.~Sukthankar, and L.~Fei-Fei,
  ``Large-scale video classification with convolutional neural networks,'' in
  \emph{IEEE Conference on Computer Vision and Pattern Recognition
  (CVPR)}.\hskip 1em plus 0.5em minus 0.4em\relax IEEE, 2014, pp. 1725--1732.

\bibitem{krizhevsky2012imagenet}
A.~Krizhevsky, I.~Sutskever, and G.~E. Hinton, ``Imagenet classification with
  deep convolutional neural networks,'' in \emph{Neural Information Processing
  Systems (NIPS)}, 2012, pp. 1097--1105.

\bibitem{Szegedy15}
C.~Szegedy, W.~Liu, Y.~Jia, P.~Sermanet, S.~Reed, D.~Anguelov, D.~Erhan,
  V.~Vanhoucke, and A.~Rabinovich, ``Going deeper with convolutions,'' in
  \emph{IEEE Conference on Computer Vision and Pattern Recognition (CVPR)},
  2015.

\bibitem{donahue2014long}
J.~Donahue, L.~Anne~Hendricks, S.~Guadarrama, M.~Rohrbach, S.~Venugopalan,
  K.~Saenko, and T.~Darrell, ``Long-term recurrent convolutional networks for
  visual recognition and description,'' in \emph{IEEE Conference on Computer
  Vision and Pattern Recognition (CVPR)}, 2015, pp. 2625--2634.

\bibitem{graves2014towards}
A.~Graves and N.~Jaitly, ``Towards end-to-end speech recognition with recurrent
  neural networks,'' in \emph{International Conference on Machine Learning
  (ICML)}, 2014, pp. 1764--1772.

\bibitem{msibrahi16deepactivity}
M.~S. Ibrahim, S.~Muralidharan, Z.~Deng, A.~Vahdat, and G.~Mori, ``A
  hierarchical deep temporal model for group activity recognition.'' in
  \emph{IEEE Conference on Computer Vision and Pattern Recognition (CVPR)},
  2016.

\bibitem{weinland2011survey}
D.~Weinland, R.~Ronfard, and E.~Boyer, ``A survey of vision-based methods for
  action representation, segmentation and recognition,'' \emph{Computer Vision
  and Image Understanding}, vol. 115, no.~2, pp. 224--241, 2011.

\bibitem{poppe2010survey}
R.~Poppe, ``A survey on vision-based human action recognition,'' \emph{Image
  and Vision Computing}, vol.~28, no.~6, pp. 976--990, 2010.

\bibitem{LanWYRM12}
T.~Lan, Y.~Wang, W.~Yang, S.~Robinovitch, and G.~Mori, ``Discriminative latent
  models for recognizing contextual group activities,'' \emph{IEEE Transactions
  on Pattern Analysis and Machine Intelligence (PAMI)}, vol.~34, no.~8, pp.
  1549--1562, 2012.

\bibitem{LanSM12}
T.~Lan, L.~Sigal, and G.~Mori, ``Social roles in hierarchical models for human
  activity recognition,'' in \emph{IEEE Conference on Computer Vision and
  Pattern Recognition (CVPR)}, 2012.

\bibitem{klaser2008spatio}
A.~Klaser, M.~Marsza{\l}ek, and C.~Schmid, ``A spatio-temporal descriptor based
  on 3d-gradients,'' in \emph{British Machine Vision Conference (BMVC))}.\hskip
  1em plus 0.5em minus 0.4em\relax British Machine Vision Association, 2008,
  pp. 275--1.

\bibitem{li2010object}
L.-J. Li, H.~Su, L.~Fei-Fei, and E.~P. Xing, ``Object bank: A high-level image
  representation for scene classification \& semantic feature sparsification,''
  in \emph{Neural Information Processing Systems (NIPS)}, 2010, pp. 1378--1386.

\bibitem{zhu2012face}
X.~Zhu and D.~Ramanan, ``Face detection, pose estimation, and landmark
  localization in the wild,'' in \emph{IEEE Conference on Computer Vision and
  Pattern Recognition (CVPR)}.\hskip 1em plus 0.5em minus 0.4em\relax IEEE,
  2012, pp. 2879--2886.

\bibitem{choiS12}
W.~Choi and S.~Savarese, ``A unified framework for multi-target tracking and
  collective activity recognition,'' in \emph{European Conference on Computer
  Vision (ECCV)}.\hskip 1em plus 0.5em minus 0.4em\relax Springer, 2012, pp.
  215--230.

\bibitem{chang2011probabilistic}
M.-C. Chang, N.~Krahnstoever, and W.~Ge, ``Probabilistic group-level motion
  analysis and scenario recognition,'' in \emph{IEEE Internation Conference on
  Computer Vision (ICCV)}.\hskip 1em plus 0.5em minus 0.4em\relax IEEE, 2011,
  pp. 747--754.

\bibitem{VasconZCHPM14}
S.~Vascon, E.~Zemene, M.~Cristani, H.~Hung, M.~Pelillo, and V.~Murino, ``A
  game-theoretic probabilistic approach for detecting conversational groups,''
  in \emph{12th Asian Conf. on Computer Vision (ACCV)}, 2014.

\bibitem{choi2011learning}
W.~Choi, K.~Shahid, and S.~Savarese, ``Learning context for collective activity
  recognition,'' in \emph{IEEE Conference on Computer Vision and Pattern
  Recognition (CVPR)}.\hskip 1em plus 0.5em minus 0.4em\relax IEEE, 2011, pp.
  3273--3280.

\bibitem{ShuXRTZ15}
T.~Shu, D.~Xie, B.~Rothrock, S.~Todorovic, and S.-C. Zhu, ``Joint inference of
  groups, events and human roles in aerial videos,'' in \emph{IEEE Conference
  on Computer Vision and Pattern Recognition (CVPR)}, 2015.

\bibitem{IntilleB01}
S.~S. Intille and A.~Bobick, ``Recognizing planned, multiperson action,''
  \emph{Computer Vision and Image Understanding (CVIU)}, vol.~81, pp. 414--445,
  2001.

\bibitem{nillius2006multi}
P.~Nillius, J.~Sullivan, and S.~Carlsson, ``Multi-target tracking-linking
  identities using bayesian network inference,'' in \emph{IEEE Conference on
  Computer Vision and Pattern Recognition (CVPR)}, 2006.

\bibitem{morariu11eventstructure}
V.~I. Morariu and L.~S. Davis, ``Multi-agent event recognition in structured
  scenarios.'' in \emph{IEEE Conference on Computer Vision and Pattern
  Recognition (CVPR)}, 2011.

\bibitem{soomro2015tracking}
K.~Soomro, S.~Khokhar, and M.~Shah, ``Tracking when the camera looks away,'' in
  \emph{IEEE International Conference on Computer Vision (ICCV) Workshops},
  2015, pp. 25--33.

\bibitem{Bo_2013_CVPR_Workshops}
Y.~Bo and H.~Jiang, ``Scale and rotation invariant approach to tracking human
  body part regions in videos,'' in \emph{IEEE Conference on Computer Vision
  and Pattern Recognition (CVPR) Workshops}, June 2013.

\bibitem{Turchini_2015_ICCV_Workshops}
F.~Turchini, L.~Seidenari, and A.~Del~Bimbo, ``Understanding sport activities
  from correspondences of clustered trajectories,'' in \emph{IEEE International
  Conference on Computer Vision (ICCV) Workshops}, December 2015.

\bibitem{KwakHH13}
S.~Kwak, B.~Han, and J.~H. Han, ``Multi-agent event detection: Localization and
  role assignment,'' in \emph{IEEE Conference on Computer Vision and Pattern
  Recognition (CVPR)}, 2013.

\bibitem{Wei_2015_ICCV_Workshops}
X.~Wei, L.~Sha, P.~Lucey, P.~Carr, S.~Sridharan, and I.~Matthews, ``Predicting
  ball ownership in basketball from a monocular view using only player
  trajectories,'' in \emph{IEEE International Conference on Computer Vision
  (ICCV) Workshops}, December 2015.

\bibitem{siddiquie2009recognizing}
B.~Siddiquie, Y.~Yacoob, and L.~Davis, ``Recognizing plays in american football
  videos,'' Technical report, University of Maryland, Tech. Rep., 2009.

\bibitem{bialkowski2013recognising}
A.~Bialkowski, P.~Lucey, P.~Carr, S.~Denman, I.~Matthews, and S.~Sridharan,
  ``Recognising team activities from noisy data,'' in \emph{IEEE Conference on
  Computer Vision and Pattern Recognition (CVPR) Workshops}, 2013, pp.
  984--990.

\bibitem{Atmosukarto_2013_CVPR_Workshops}
I.~Atmosukarto, B.~Ghanem, S.~Ahuja, K.~Muthuswamy, and N.~Ahuja, ``Automatic
  recognition of offensive team formation in american football plays,'' in
  \emph{IEEE Conference on Computer Vision and Pattern Recognition (CVPR)
  Workshops}, June 2013.

\bibitem{direkoǧluO12}
C.~Direkoglu and N.~E. O'Connor, ``Team activity recognition in sports,'' in
  \emph{European Conference on Computer Vision (ECCV)}.\hskip 1em plus 0.5em
  minus 0.4em\relax Springer, 2012, pp. 69--83.

\bibitem{SwearsHJB14}
E.~Swears, A.~Hoogs, Q.~Ji, and K.~Boyer, ``Complex activity recognition using
  granger constrained dbn (gcdbn) in sports and surveillance video,'' in
  \emph{IEEE Conference on Computer Vision and Pattern Recognition (CVPR)},
  June 2014.

\bibitem{Gade_2013_CVPR_Workshops}
R.~Gade and T.~B. Moeslund, ``Sports type classification using signature
  heatmaps,'' in \emph{IEEE Conference on Computer Vision and Pattern
  Recognition (CVPR) Workshops}, June 2013.

\bibitem{simonyan2014very}
K.~Simonyan and A.~Zisserman, ``Very deep convolutional networks for
  large-scale image recognition,'' \emph{CoRR}, vol. abs/1409.1556, 2014.

\bibitem{hochreiter1997long}
S.~Hochreiter and J.~Schmidhuber, ``Long short-term memory,'' \emph{Neural
  computation}, vol.~9, no.~8, pp. 1735--1780, 1997.

\bibitem{ng2015beyond}
J.~Y.-H. Ng, M.~Hausknecht, S.~Vijayanarasimhan, O.~Vinyals, R.~Monga, and
  G.~Toderici, ``Beyond short snippets: Deep networks for video
  classification,'' \emph{IEEE Conference on Computer Vision and Pattern
  Recognition (CVPR)}, 2015.

\bibitem{venugopalan2014translating}
S.~Venugopalan, H.~Xu, J.~Donahue, M.~Rohrbach, R.~Mooney, and K.~Saenko,
  ``Translating videos to natural language using deep recurrent neural
  networks,'' in \emph{North American Chapter of the Association for
  Computational Linguistics}, 2015.

\bibitem{karpathy2014deep}
A.~Karpathy and L.~Fei-Fei, ``Deep visual-semantic alignments for generating
  image descriptions,'' \emph{IEEE Conference on Computer Vision and Pattern
  Recognition (CVPR)}, 2015.

\bibitem{schuster1997bidirectional}
M.~Schuster and K.~K. Paliwal, ``Bidirectional recurrent neural networks,''
  \emph{IEEE Transcations on Signal Processing}, vol.~45, no.~11, pp.
  2673--2681, 1997.

\bibitem{NIPS2014_5573}
J.~J. Tompson, A.~Jain, Y.~LeCun, and C.~Bregler, ``Joint training of a
  convolutional network and a graphical model for human pose estimation,'' in
  \emph{Advances in Neural Information Processing Systems 27}, Z.~Ghahramani,
  M.~Welling, C.~Cortes, N.~Lawrence, and K.~Weinberger, Eds.\hskip 1em plus
  0.5em minus 0.4em\relax Curran Associates, Inc., 2014, pp. 1799--1807.

\bibitem{Zheng15}
S.~Zheng, S.~Jayasumana, B.~Romera-Paredes, V.~Vineet, Z.~Su, D.~Du, C.~Huang,
  and P.~H.~S. Torr, ``Conditional random fields as recurrent neural
  networks,'' in \emph{IEEE International Conference on Computer Vision
  (ICCV)}, 2015.

\bibitem{schwing2015fully}
A.~G. Schwing and R.~Urtasun, ``Fully connected deep structured networks,''
  \emph{arXiv preprint arXiv:1503.02351}, 2015.

\bibitem{DengZCLMRM15}
Z.~Deng, M.~Zhai, L.~Chen, Y.~Liu, S.~Muralidharan, M.~Roshtkhari, and G.~Mori,
  ``Deep structured models for group activity recognition,'' in \emph{British
  Machine Vision Conference (BMVC)}, 2015.

\bibitem{yeung2015every}
S.~Yeung, O.~Russakovsky, N.~Jin, M.~Andriluka, G.~Mori, and L.~Fei-Fei,
  ``Every moment counts: Dense detailed labeling of actions in complex
  videos,'' \emph{arXiv preprint arXiv:1507.05738}, 2015.

\bibitem{RamanathanHAGMF16}
V.~Ramanathan, J.~Huang, S.~Abu-El-Haija, A.~Gorban, K.~Murphy, and L.~Fei-Fei,
  ``Detecting events and key actors in multi-person videos,'' in \emph{IEEE
  Conference on Computer Vision and Pattern Recognition (CVPR)}, 2016.

\bibitem{soomro2012ucf101}
K.~Soomro, A.~R. Zamir, and M.~Shah, ``Ucf101: A dataset of 101 human actions
  classes from videos in the wild,'' \emph{arXiv preprint arXiv:1212.0402},
  2012.

\bibitem{kuehne2011hmdb}
H.~Kuehne, H.~Jhuang, E.~Garrote, T.~Poggio, and T.~Serre, ``Hmdb: a large
  video database for human motion recognition,'' in \emph{IEEE International
  Conference on Computer Vision (ICCV)}.\hskip 1em plus 0.5em minus 0.4em\relax
  IEEE, 2011, pp. 2556--2563.

\bibitem{2015trecvidover}
P.~Over, G.~Awad, M.~Michel, J.~Fiscus, W.~Kraaij, A.~F. Smeaton, G.~Quéenot,
  and R.~Ordelman, ``Trecvid 2015 -- an overview of the goals, tasks, data,
  evaluation mechanisms and metrics,'' in \emph{Proceedings of TRECVID
  2015}.\hskip 1em plus 0.5em minus 0.4em\relax NIST, USA, 2015.

\bibitem{UT-Interaction-Data}
M.~S. Ryoo and J.~K. Aggarwal, ``{UT}-{I}nteraction {D}ataset, {ICPR} contest
  on {S}emantic {D}escription of {H}uman {A}ctivities ({SDHA}),''
  http://cvrc.ece.utexas.edu/SDHA2010/Human\_Interaction.html, 2010.

\bibitem{OhVIRAT11}
S.~Oh, A.~Hoogs, A.~Perera, N.~Cuntoor, C.-C. Chen, J.~T. Lee, S.~Mukherjee,
  J.~Aggarwal, H.~Lee, L.~Davis, E.~Swears, X.~Wang, Q.~Ji, K.~Reddy, M.~Shah,
  C.~Vondrick, H.~Pirsiavash, D.~Ramanan, J.~Yuen, A.~Torralba, B.~Song,
  A.~Fong, A.~Roy-Chowdhury, and M.~Desai, ``A large-scale benchmark dataset
  for event recognition in surveillance video,'' in \emph{IEEE Conference on
  Computer Vision and Pattern Recognition (CVPR)}, 2011.

\bibitem{amer2012cost}
M.~R. Amer, D.~Xie, M.~Zhao, S.~Todorovic, and S.-C. Zhu, ``Cost-sensitive
  top-down/bottom-up inference for multiscale activity recognition,'' in
  \emph{European Conference on Computer Vision (ECCV)}.\hskip 1em plus 0.5em
  minus 0.4em\relax Springer, 2012, pp. 187--200.

\bibitem{ConigliaroRSBCSC15}
D.~Conigliaro, P.~Rota, F.~Setti, C.~Bassetti, N.~Conci, N.~Sebe, and
  M.~Cristani, ``The s-hock dataset: Analyzing crowds at the stadium,'' in
  \emph{IEEE Conference on Computer Vision and Pattern Recognition (CVPR)},
  2015.

\bibitem{lazebnik2006beyond}
S.~Lazebnik, C.~Schmid, and J.~Ponce, ``Beyond bags of features: Spatial
  pyramid matching for recognizing natural scene categories,'' in \emph{IEEE
  Conference on Computer Vision and Pattern Recognition (CVPR)}, 2006.

\bibitem{Caffe}
Y.~Jia, ``Caffe: An open source convolutional architecture or fast feature
  embedding,'' 2013, http://caffe.berkeleyvision.org/.

\bibitem{Danelljan14_tracker}
M.~Danelljan, G.~Häger, F.~Shahbaz~Khan, and M.~Felsberg, ``Accurate scale
  estimation for robust visual tracking,'' in \emph{British Machine Vision
  Conference (BMVC)}, 2014.

\bibitem{dlib09}
D.~E. King, ``Dlib-ml: A machine learning toolkit,'' \emph{Journal of Machine
  Learning Research}, vol.~10, pp. 1755--1758, 2009.

\bibitem{hajimirsadeghi2015visual}
H.~Hajimirsadeghi, W.~Yan, A.~Vahdat, and G.~Mori, ``Visual recognition by
  counting instances: A multi-instance cardinality potential kernel,''
  \emph{IEEE Conference on Computer Vision and Pattern Recognition (CVPR)},
  2015.

\bibitem{Wang2013}
\BIBentryALTinterwordspacing
H.~Wang and C.~Schmid, ``Action recognition with improved trajectories,'' in
  \emph{IEEE International Conference on Computer Vision (ICCV)}, Sydney,
  Australia, 2013. [Online]. Available: \url{http://hal.inria.fr/hal-00873267}
\BIBentrySTDinterwordspacing

\end{thebibliography}

\begin{IEEEbiography}{Mostafa S. Ibrahim}
was born in Egypt and received MSc degree in Computer Science from Cairo University in 2012, and B.Sc. in 2008 from same university. His research areas of interest include image classification, object detection, object tracking and human activity recognition. Mostafa is currently doing PhD degree in Computer Science at Simon Fraser University under supervision of Prof Greg Mori. Moreover, Mostafa worked as software engineer for several years and has wide interest in competitive programming. He was a 2011 ACM ICPC world finalist and played later several roles such as Chief Judge, Problem Setter and Coach.
\end{IEEEbiography}

\begin{IEEEbiography}{Srikanth Muralidharan}
received his BTech in Electrical Engineering from Indian Institute of Technology Jodhpur in 2014. After Graduating from IIT Jodhpur, he is now a Masters Student at Simon Fraser University working with Prof. Mori.  His research interest is in Machine learning, with applications related to video analysis. 
\end{IEEEbiography}

\begin{IEEEbiography}{Zhiwei Deng}
Zhiwei Deng is currently a Ph.D. candidate in the School of Computing Science at Simon Fraser University, Canada. He received his M.Sc. from the same university in 2015, and his B.Eng from Beijing University of Posts and Telecommunications, China in 2013. His research interests are in the area of computer vision and deep learning, with a focus on understanding of human actions, group activities with deep structured models.
\end{IEEEbiography}

\begin{IEEEbiography}{Arash Vahdat}
joined the School of Computing Science, Simon Fraser University as a research faculty in 2015. His research interest relies in the area of weakly supervised learning, structured prediction and deep learning for computer vision applications and he publishes his research in the top computer vision conferences (e.g. CVPR, ICCV, ECCV, NIPS). As a University Research Associate, he taught several courses for SFU's Professional Master's program in Big Data. In 2014, he worked at Google's image search team in Mountain View, CA as an intern. Dr. Vahdat obtained his PhD and MSc in Computer Science from Simon Fraser University and he holds a BSc in Computer Engineering from Sharif University of Technology.
\end{IEEEbiography}

\begin{IEEEbiography}{Greg Mori}
was born in Vancouver and grew up in Richmond, BC. He received the Ph.D. degree in Computer Science from the University of California, Berkeley in 2004. He received an Hon. B.Sc. in Computer Science and Mathematics with High Distinction from the University of Toronto in 1999. He spent one year (1997-1998) as an intern at Advanced Telecommunications Research (ATR) in Kyoto, Japan. After graduating from Berkeley, he returned home to Vancouver and is currently a professor in the School of Computing Science at Simon Fraser University. Dr. Mori’s research interests are in computer vision, and include object recognition, human activity recognition, human body pose estimation. He serves on the program committee of major computer vision conferences (CVPR, ECCV, ICCV), and was the program co-chair of the Canadian Conference on Computer and Robot Vision (CRV) in 2006 and 2007. Dr. Mori received the Excellence in Undergraduate Teaching Award from the SFU Computing Science Student Society in 2006. Dr. Mori received the Canadian Image Processing and Pattern Recognition Society (CIPPRS) Award for Research Excellence and Service in 2008. Dr. Mori received NSERC Discovery Accelerator Supplement awards in 2008 and 2016.
\end{IEEEbiography}

\end{document}